%% file: neurips_2026.tex
\documentclass{article}

\PassOptionsToPackage{numbers, compress}{natbib}

 \usepackage[main, final]{neurips_2026}

\usepackage[utf8]{inputenc} 
\usepackage[T1]{fontenc}    
\usepackage{hyperref}       
\usepackage{url}            
\usepackage{booktabs}       
\usepackage{amsfonts}       
\usepackage{nicefrac}       
\usepackage{microtype}      
\usepackage{xcolor}         

\usepackage{amsmath}
\usepackage{amssymb}
\usepackage{mathtools}
\usepackage{amsthm}
\renewcommand{\cite}{\citep}
\usepackage[T1]{fontenc}    
\usepackage{hyperref}       
\usepackage{url}            
\usepackage{booktabs}       
\usepackage{amsfonts}       
\usepackage{nicefrac}       
\usepackage{microtype}      
\usepackage{xcolor}         
\definecolor{citecolor}{HTML}{0071BC}
\definecolor{linkcolor}{HTML}{D32F2F}
\definecolor{cellcolor}{HTML}{E3F2FD}
\definecolor{red}{HTML}{D32F2F}
\definecolor{magenta}{HTML}{D81B60}

\usepackage{svg}
\svgsetup{inkscapearea=page}
\usepackage{amsmath}
\usepackage{amssymb}
\usepackage{mathtools}
\usepackage{amsthm}
\usepackage{amsmath, amssymb, amsthm}
\usepackage{amsfonts}
\usepackage[capitalize,noabbrev]{cleveref}
\newtheorem{assumption}{\textbf{Assumption}}

\newtheorem{lemma}{\textbf{Lemma}}
\newtheorem{theorem}{\textbf{Theorem}}

\newtheorem{remark}{\textbf{Remark}}

\usepackage{pgfplots}
\pgfplotsset{compat = newest}
\usepackage{multirow}
\usepackage{enumitem}
\usepackage{caption}
\usepackage{fancybox}
\usepackage{framed}
\usepackage{tcolorbox}
\usepackage{subcaption}
\usepackage{algorithm}
\usepackage{mathrsfs}
\usepackage{mathtools}
\usepackage{algorithmic}
\usepackage{tcolorbox}
\usepackage{listings}
\usepackage{makecell}
\usepackage{colortbl}
\usepackage{color}
\usepackage{wrapfig}
\usepackage{cancel}
\usepackage{soul,xcolor}
\usepackage{pifont}
\usepackage{tikz}
\usetikzlibrary{shapes, positioning, arrows.meta, calc, decorations.pathmorphing, quotes}
\tcbuselibrary{breakable}
\usepackage{hyperref}
\usepackage{makecell}
\usepackage{url}
\hypersetup{colorlinks=true, linkcolor=linkcolor, citecolor=citecolor,urlcolor=magenta}
\usepackage{booktabs}
\usepackage{hyperref}
\usepackage{url}
\usepackage{graphicx}
\usepackage{subcaption}
\usepackage{amsthm}

\usepackage{enumitem}
\setlist[itemize]{leftmargin=2em}
\setlist[enumerate]{leftmargin=2em}

\DeclareMathOperator{\clip}{clip}

\usepackage[textsize=tiny]{todonotes}
\usepackage{microtype}
\usepackage{graphicx}
\usepackage{booktabs} 
\title{SCOPE-RL: Stable and Quantitative Control of Policy Entropy in RL Post-Training}

\author{%
\begin{tabular}{c}
Chen Wang$^{1,2,*}$ \quad
Zhaochun Li$^{2,3,*}$ \quad
Jionghao Bai$^{2,4,*}$ \\
Hexuan Deng$^{2,5}$ \quad
Guanting Dong$^{6}$ \quad
Ge Lan$^{7,\dagger}$ \quad
Yue Wang$^{2,\dagger}$ \\[0.8em]
\small
\normalfont $^{1}$College of Elite Engineers, Nankai University \quad
\normalfont $^{2}$Zhongguancun Academy \\
\small
\normalfont $^{3}$Beijing Institute of Technology \quad
\normalfont $^{4}$Zhejiang University \quad
\normalfont $^{5}$Harbin Institute of Technology \\
\small
\normalfont $^{6}$Renmin University of China\quad
\normalfont $^{7}$College of Software, Nankai University
\end{tabular}
}

\begin{document}

\maketitle

\begin{abstract}
Reinforcement learning (RL) is a key paradigm for post-training large language models (LLMs), but the widely used Group Relative Policy Optimization (GRPO) often suffers from entropy collapse: exploration quickly disappears, policies converge prematurely, and sample diversity declines, ultimately harming training effectiveness. Existing remedies, including entropy bonuses and clip-based methods, rarely keep entropy within a stable exploration regime and often introduce oscillatory entropy or reward degradation. In this work, we identify a previously overlooked asymmetry in entropy dynamics: under high-temperature sampling, positive and negative samples have opposite effects on policy entropy. Specifically, high-temperature positive samples promote entropy growth, whereas negative samples suppress it. We provide a qualitative theoretical explanation for this phenomenon: when entropy decreases during policy updates, its derivative with respect to temperature is strictly positive under positive sample updates, indicating that high-temperature positive samples can counteract entropy decay, thereby slowing entropy collapse and potentially reversing it. Motivated by this insight, we propose SCOPE-RL, a stable entropy control framework through a regularization term constructed from temperature-adaptive positive samples. Extensive experiments show that SCOPE-RL consistently outperforms strong RL baselines on both Pass@1 and Pass@k. Our results provide evidence that escaping entropy collapse can improve reasoning performance, while our quantitative empirical analysis further reveals a non-monotonic entropy–performance relationship, with an optimal level of exploration for RL post-training in reasoning LLMs. Our code is available at \url{https://github.com/597358816/SCOPE-RL}.
\end{abstract}

\section{Introduction}

\begingroup
\renewcommand{\thefootnote}{}
\footnotetext{$^*$: Contribute equally.}
\footnotetext{$^\dagger$ Correspondence to Ge Lan, email: \texttt{lange@nankai.edu.cn}.}
\footnotetext{$^\dagger$ Correspondence to Yue Wang, email: \texttt{yuewang@bza.edu.cn}.}
\endgroup

\begin{figure}[h]
	\centering
	\includegraphics[width=\linewidth]{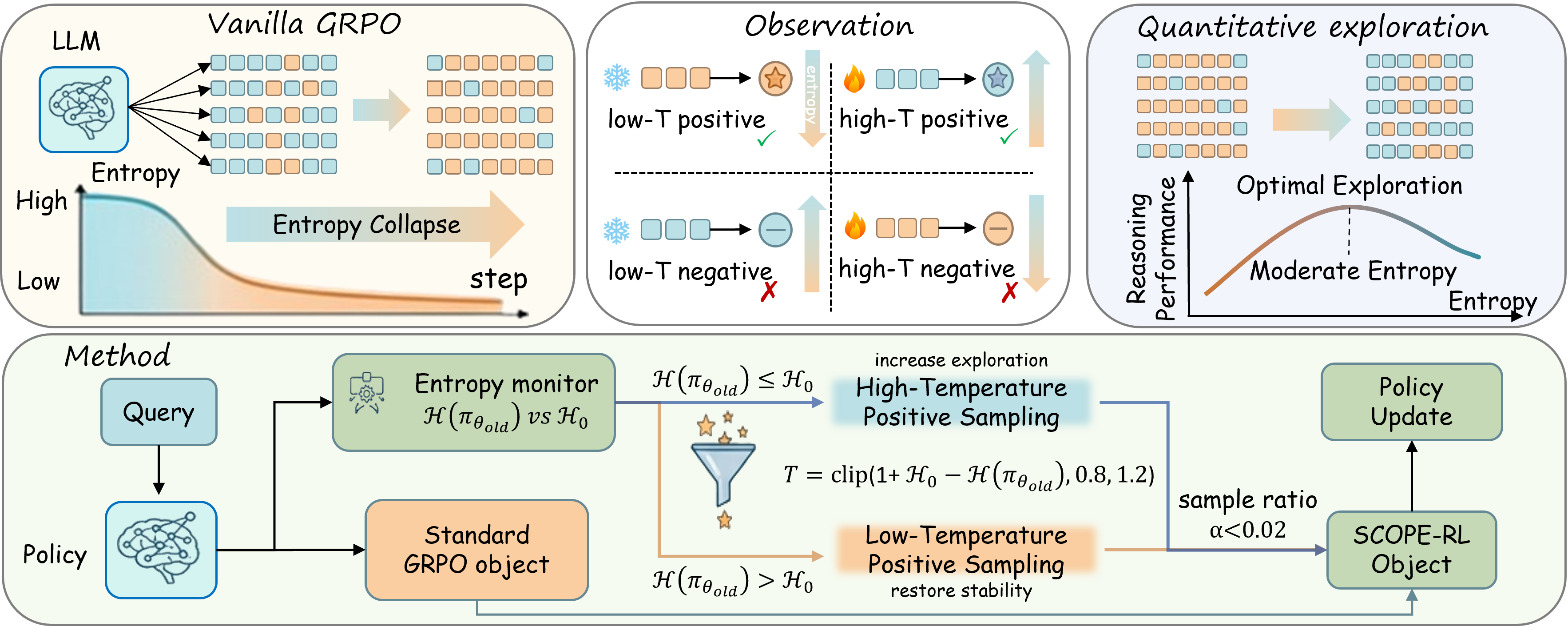}
	\caption{The overview of SCOPE-RL.}
	\label{fig:main}
	\vspace{-12pt}
\end{figure}

\textbf{Reinforcement learning} (RL) has become a central paradigm for improving the reasoning capabilities of large language models (LLMs) \cite{glm2024chat, touvron2023llama, schulman2017proximal, rafailov2023direct, zhong2024dpo, wang2024comprehensive}. A key ingredient of RL is \textbf{exploration}, which requires the policy to generate sufficiently diverse and informative samples during training \cite{sutton1998reinforcement, auer2002finite, strehl2008analysis, kolter2009near}. In reasoning LLMs, exploration is closely related to \textbf{entropy}, a measure of policy uncertainty and output diversity \cite{schulman2017equivalence, haarnoja2018soft, nachum2017bridging}. Among existing methods, Group Relative Policy Optimization (GRPO) has become a widely used backbone due to its efficiency and scalability \cite{shao2024deepseekmath, liu2024deepseek, guo2025deepseek}. However, GRPO suffers from a well-known failure mode: \textbf{entropy collapse}, where policy entropy steadily decreases, outputs become increasingly homogeneous, and exploration vanishes prematurely \cite{yu2025dapo, li2025disco, zhang2025edge}. As a result, RL often fails to discover genuinely diverse reasoning strategies and may mainly sharpen behaviors already latent in the base model \cite{yue2025does}. These findings point to a central challenge in RL post-training for reasoning LLMs: effective improvement requires restoring exploration, and restoring exploration requires overcoming entropy collapse \cite{cui2025entropy}.

Although entropy collapse in GRPO has been repeatedly recognized, it remains fundamentally unresolved. Existing attempts to alleviate this issue mainly follow two directions: entropy bonuses \cite{cui2025entropy, hou2025advancing, shen2025entropy, cheng2025reasoning} and clip-based methods \cite{yu2025dapo, cui2025entropy,  chen2025minimax, su2025gppo}. However, both often introduce unstable training dynamics that undermine training effectiveness. In practice, entropy may oscillate between collapse and explosion, accompanied by irreversible performance degradation, rather than stabilizing at an exploration regime that is beneficial for learning. This instability often offsets the benefits brought by exploration and can even harm training. Therefore, there is still a lack of a stable entropy control mechanism that enables quantitative analysis of the entropy–performance relationship and helps identify the optimal degree of exploration in the exploration–exploitation trade-off.

To address this limitation, we revisit entropy regulation in RL post-training through the lens of sample-level entropy dynamics. In this work, we identify a previously overlooked asymmetry in entropy dynamics: positive and negative samples exert opposite effects on policy entropy, and this effect reverses across temperature regimes. Under high-temperature sampling, positive samples increase entropy, whereas negative samples accelerate entropy collapse; under low-temperature sampling, the trend reverses. We provide a qualitative theoretical explanation for this phenomenon: when entropy decreases during policy updates, the derivative of entropy with respect to temperature is strictly positive under positive-sample updates, indicating that high-temperature positive samples can counteract entropy decay, thereby slowing entropy collapse and potentially reversing it. Motivated by this insight, we propose \textbf{SCOPE-RL} (Fig.~\ref{fig:main}), a stable and quantitative entropy control framework via a regularization term built from less than 2\% of temperature-adaptive positive samples. Through this mechanism, SCOPE-RL enables stable exploration control during RL post-training and maintains entropy within a controllable range, allowing us to quantitatively analyze the empirical relationship between entropy and training performance. 

Our contributions are summarized as follows:
\begin{itemize}
\item We identify a previously overlooked temperature--sign asymmetry in sample-level entropy dynamics during RL post-training: positive and negative samples have opposite effects on policy entropy, and the direction reverses across temperature regimes.
    
\item We provide a qualitative theoretical analysis showing that, under entropy collapse, the derivative of entropy with respect to temperature is strictly positive under positive-sample updates. Motivated by this insight, we propose SCOPE-RL, a stable entropy control framework using a temperature-adaptive positive-sample regularizer constructed from less than 2\% of samples.
    
\item We conduct extensive experiments on mathematical reasoning and knowledge-intensive benchmarks, showing that SCOPE-RL consistently outperforms strong RL baselines on both Pass@1 and Pass@$k$. On in-domain mathematical reasoning, it improves over GRPO by +3.23 on Qwen2.5-Math-7B, +2.10 on Qwen2.5-7B, and +1.96 on Qwen3-4B. Beyond Pass@1, it also surpasses both the base model and GRPO at Pass@1024 on AIME24 and AIME25.
    
\item Our quantitative empirical analysis further shows that escaping entropy collapse improves performance, while the entropy--performance relationship is non-monotonic, with an optimal level of exploration in RL post-training.
\end{itemize}

\section{Preliminary}

We consider reinforcement learning post-training with verifiable rewards (RLVR) for reasoning tasks such as mathematical problem solving and code generation. Let $q$ denote a query and $o=(o_1,\dots,o_{|o|})$ denote a response generated autoregressively by a language model policy. We write the standard policy and its temperature-scaled variant as
\begin{equation}
\pi_{\theta}(o \mid q)=\prod_{t=1}^{|o|}\pi_{\theta}(o_t \mid q,o_{<t}), \qquad
\pi_{\theta}^{T}(o_t \mid q,o_{<t})=
\frac{\exp(l(q,o_{<t},o_t)/T)}
{\sum_{v}\exp(l(q,o_{<t},v)/T)},
\end{equation}
where $l(q,o_{<t},o_t)$ is the logit of token $o_t$ conditioned on $(q,o_{<t})$.

\subsection{RL Post-Training and GRPO}

In RLVR, each sampled response $o$ is evaluated by a verifiable reward function $R(q,o)$, and the RL objective is to maximize the expected reward
\begin{equation}
\mathcal{J}_{\rm RL}(\theta)=
\mathbb{E}_{q \sim P(Q),\, o \sim \pi_{\theta}(O\mid q)}[R(q,o)].
\end{equation}
Among existing methods, Group Relative Policy Optimization (GRPO) is a widely used backbone for reasoning LLMs. For each query $q$, it samples a group of $G$ responses $\{o_i\}_{i=1}^{G}$ from $\pi_{\theta_{\rm old}}$ and computes the objective
\begin{equation}
\small
\mathcal{J}_{\rm GRPO}(\theta)=
\mathbb{E}_{q\sim P(Q),\,\{o_i\}_{i=1}^{G}\sim \pi_{\theta_{\rm old}}}
\left[
\frac{1}{G}\sum_{i=1}^{G}\frac{1}{|o_i|}\sum_{t=1}^{|o_i|}
\min\!\Big(
r_{i,t}(\theta)\hat{A}_i,\,
\mathrm{clip}\!\big(r_{i,t}(\theta),1-\epsilon_{\text{low}},1+\epsilon_{\text{high}}\big)\hat{A}_i
\Big)
\right],
\end{equation}
where $\epsilon_{\text{low}} = \epsilon_{\text{high}}=0.2$ and
\begin{equation}
r_{i,t}(\theta)=
\frac{\pi_{\theta}(o_{i,t}\mid q,o_{i,<t})}
{\pi_{\theta_{\rm old}}(o_{i,t}\mid q,o_{i,<t})}, \qquad
\hat{A}_i=
\frac{
R(q,o_i)-\mathrm{mean}(\{R(q,o_j)\}_{j=1}^{G})
}{
\mathrm{std}(\{R(q,o_j)\}_{j=1}^{G})
}.
\end{equation}
By using group-normalized relative advantages, GRPO reduces variance and improves training stability. However, its policy entropy often decreases monotonically during training, leading to entropy collapse and premature loss of exploration.

\begin{figure}[t]
	\centering
	
	\begin{minipage}{0.49\linewidth}
		\centering
		\includegraphics[width=0.49\linewidth]{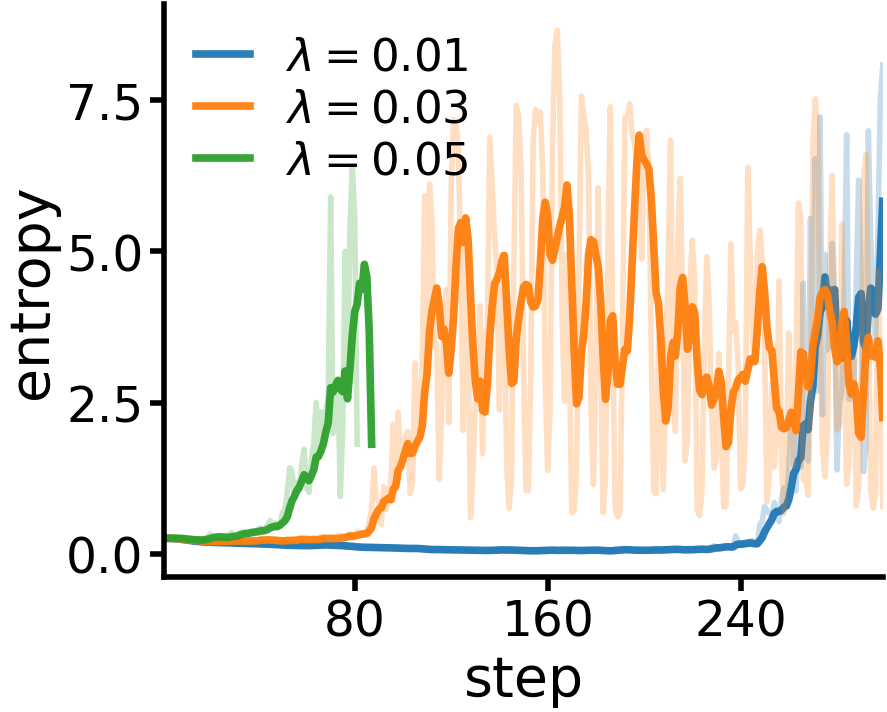}
		\includegraphics[width=0.49\linewidth]{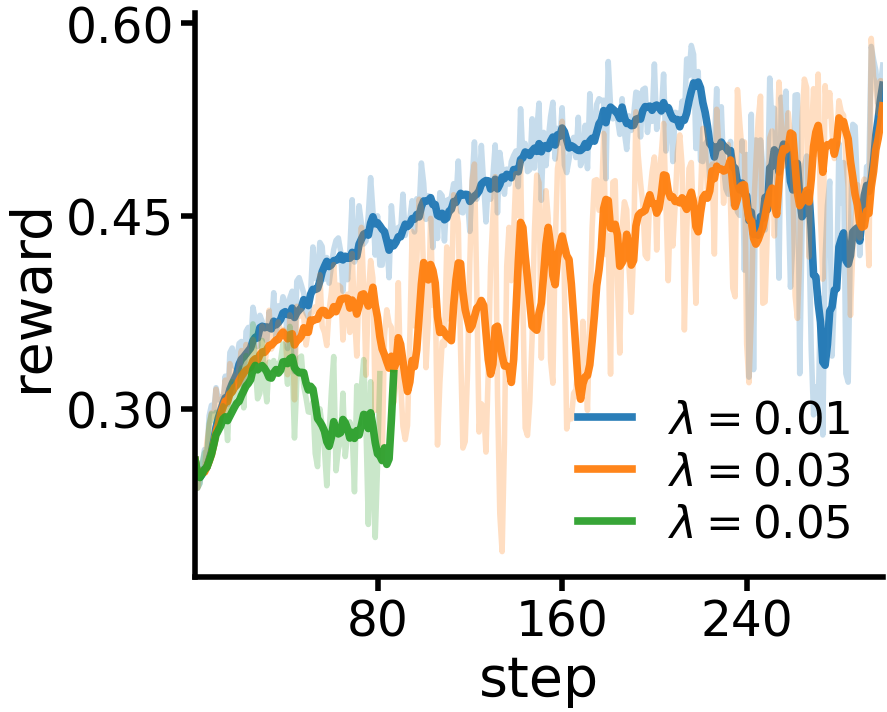}
		\subcaption{Entropy regularization}
	\end{minipage}
	\hfill
	\begin{minipage}{0.49\linewidth}
		\centering
		\includegraphics[width=0.49\linewidth]{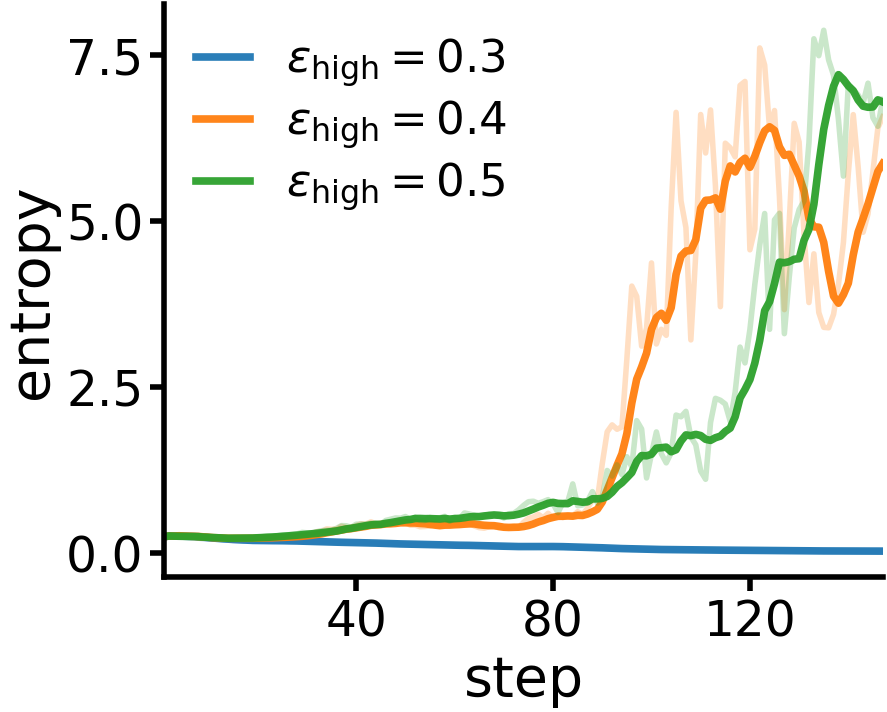}
		\includegraphics[width=0.49\linewidth]{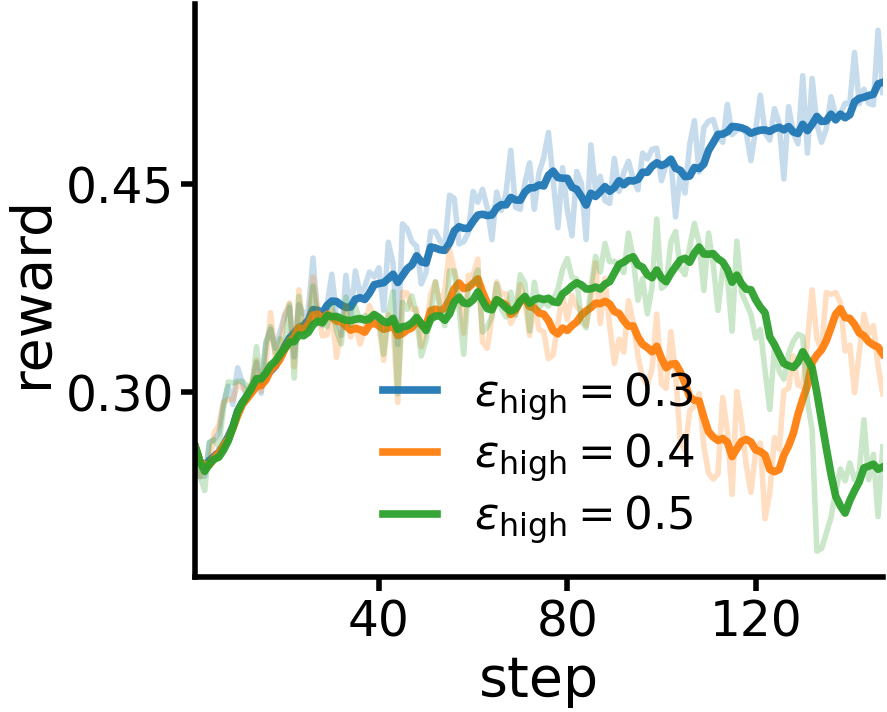}
		\subcaption{Clip higher}
	\end{minipage}
	
	\caption{Training dynamics of Qwen2.5-Math-7B on the DAPO-17k dataset under different settings. Although both methods can induce entropy growth, the increase is often accompanied by unstable training dynamics and noticeable reward degradation.}
	\label{fig:entropy_instability}
\end{figure}

\subsection{Related Methods for Entropy}

Existing remedies mainly fall into two categories: entropy regularization and clip-higher methods.

\paragraph{Entropy Regularization.}
The first category follows the principle of maximum-entropy reinforcement learning, which augments the reward objective with an entropy term to encourage exploration. For autoregressive language models, the token-level entropy and entropy-regularized objective are
\begin{equation}
\begin{split}
&\mathcal{H}(\pi_{\theta})=
\mathbb{E}_{q\sim P(Q),\,o\sim\pi_\theta(O\mid q)}
\left[
-\frac{1}{|o|}\sum_{t=1}^{|o|}
\sum_{v}\pi_{\theta}(v\mid q,o_{<t})\log\pi_{\theta}(v\mid q,o_{<t})
\right], \\
&\mathcal{J}_{\rm MaxEnt}(\theta)=
\mathcal{J}_{\rm RL}(\theta)+\lambda\,\mathcal{H}(\pi_{\theta}).
\end{split}
\end{equation}
where $\lambda>0$ controls the exploration strength. In GRPO-based training, such methods typically add entropy terms into the objective, rewards, or advantages. They are principled, but often introduce optimization bias and unstable entropy dynamics.

\paragraph{Clip higher.}
The second category modifies the symmetric clipping rule in GRPO to preserve diversity more aggressively, e.g., replacing $\mathrm{clip}(r,0.8,1.2)$ with an asymmetric form
\begin{equation}
\mathrm{clip}(r,1-\epsilon_{\rm low},1+\epsilon_{\rm high}), \qquad \epsilon_{\rm high}>\epsilon_{\rm low}.
\end{equation}

For instance, DAPO enlarges the clipping range by setting $\epsilon_{\rm high}=0.3$ and $\epsilon_{\rm low}=0.2$, and empirically observes an increasing trend in entropy.

However, both Entropy Regularization and clip-higher often exhibit clear instability in practice, as shown in Fig.~\ref{fig:entropy_instability} and Table~\ref{tab:main_results_math}. As entropy increases, both methods tend to suffer from unstable training dynamics and noticeable reward drops. This suggests that both categories have difficulty delivering reliable entropy control and consistently promoting effective exploration.

\section{Method}
\subsection{Observation and Analysis}
\begin{figure}[t]
\centering
\includegraphics[width=0.3\linewidth]{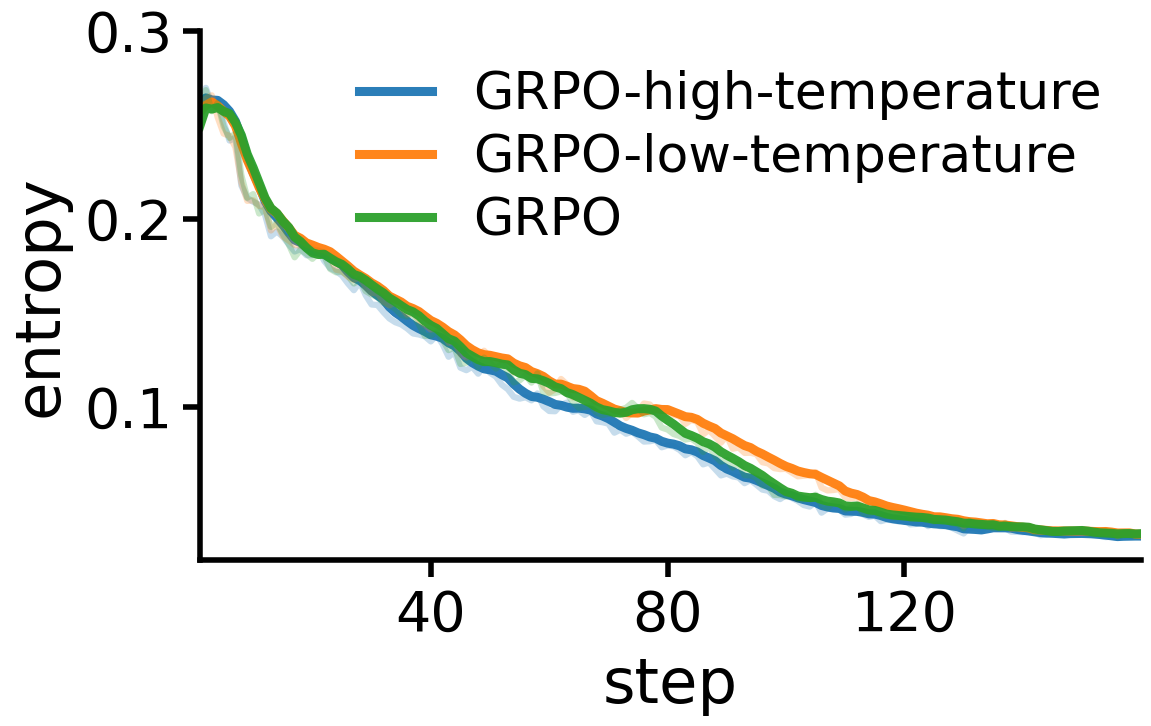}
\includegraphics[width=0.3\linewidth]{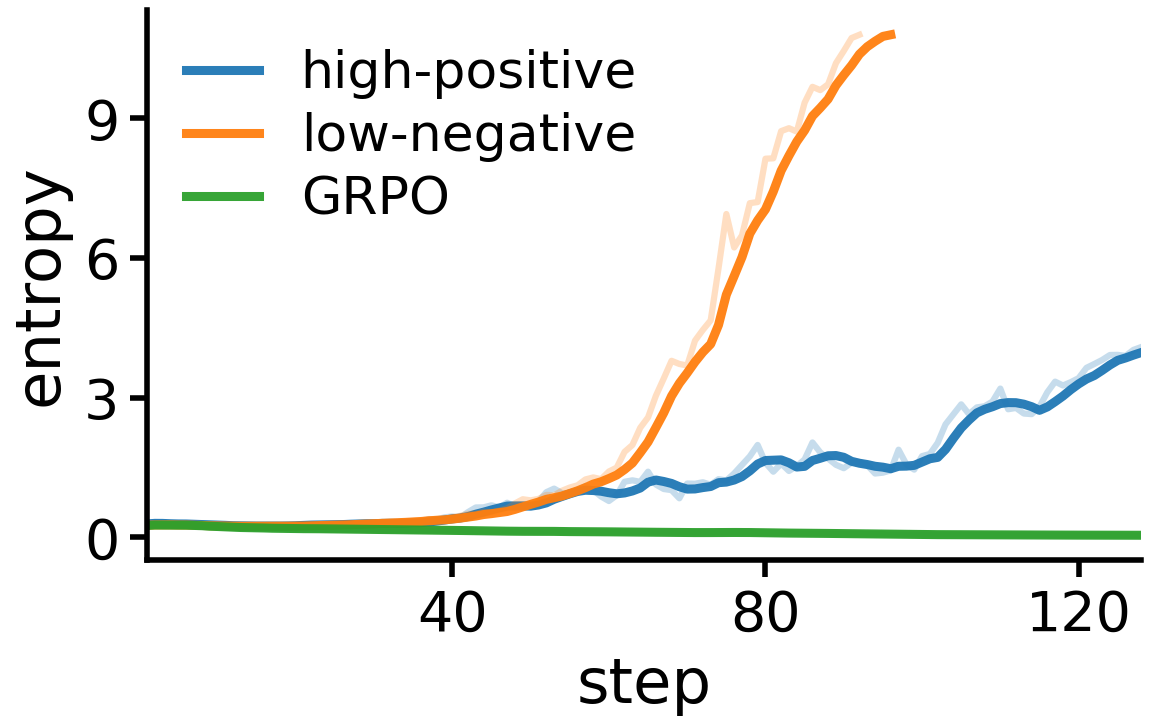}
\includegraphics[width=0.3\linewidth]{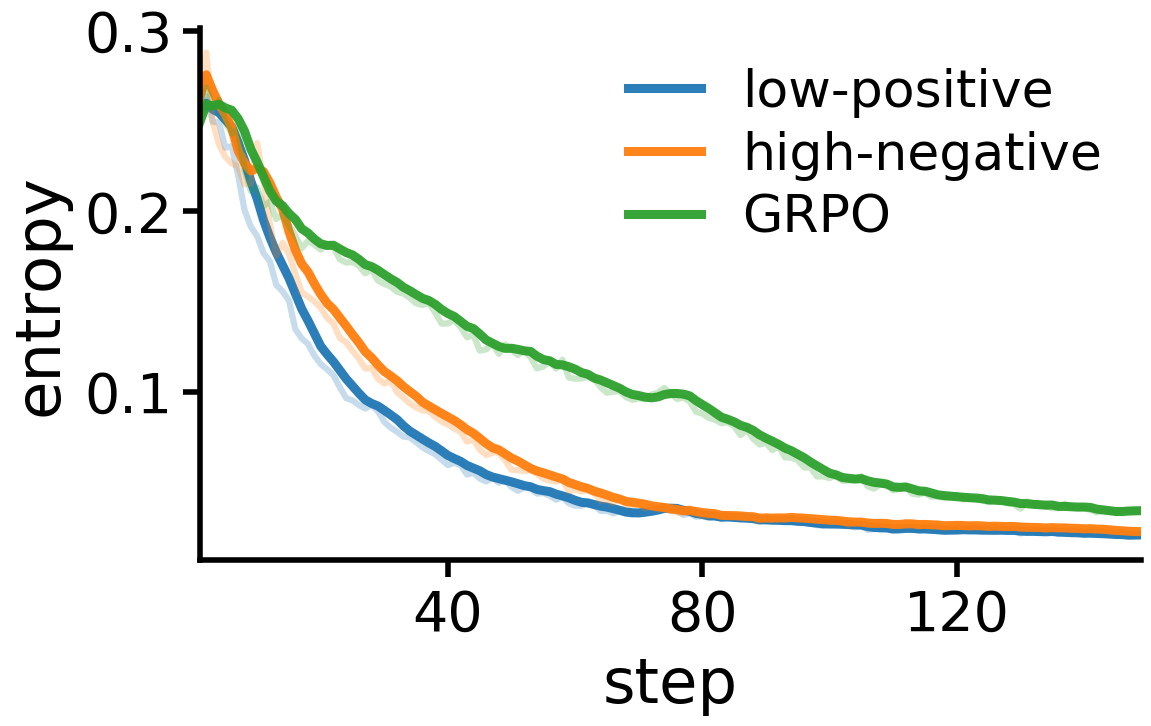} \\
\includegraphics[width=0.3\linewidth]{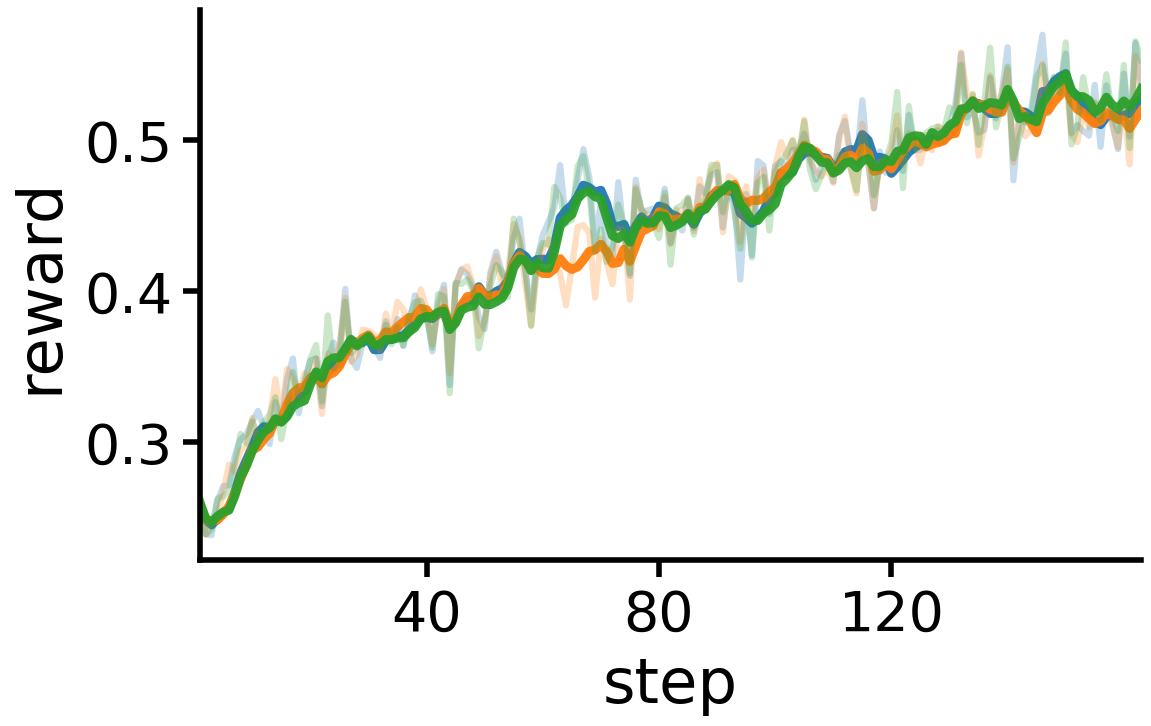}
\includegraphics[width=0.3\linewidth]{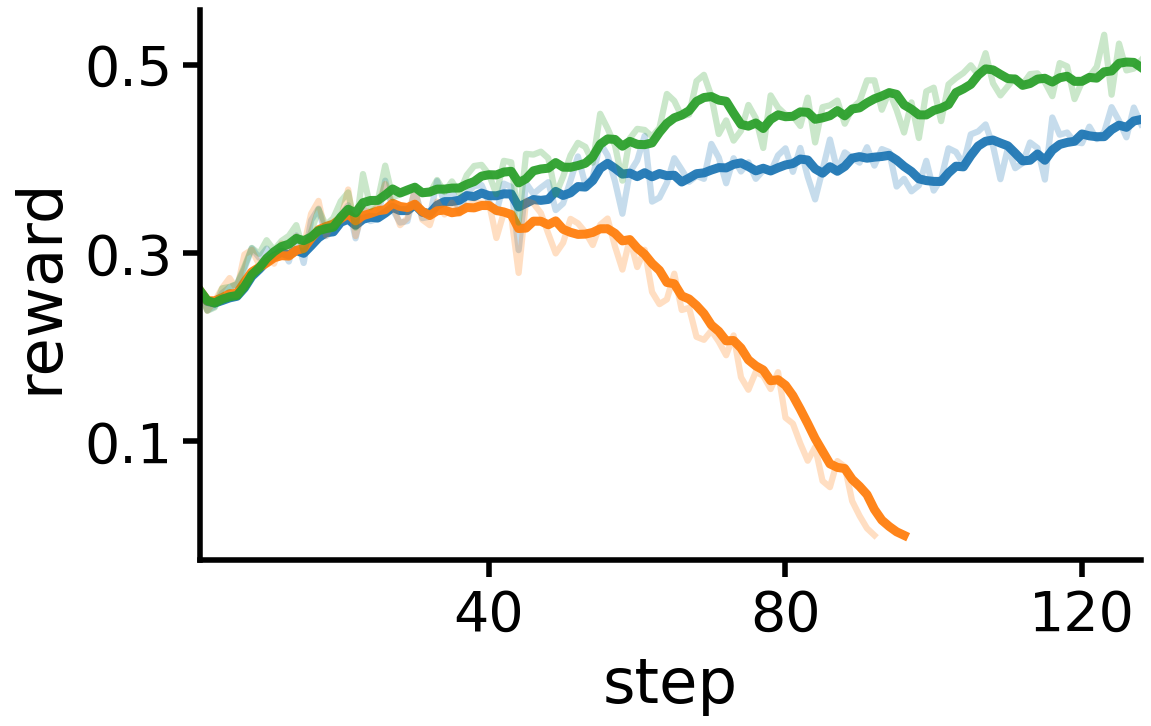}
\includegraphics[width=0.3\linewidth]{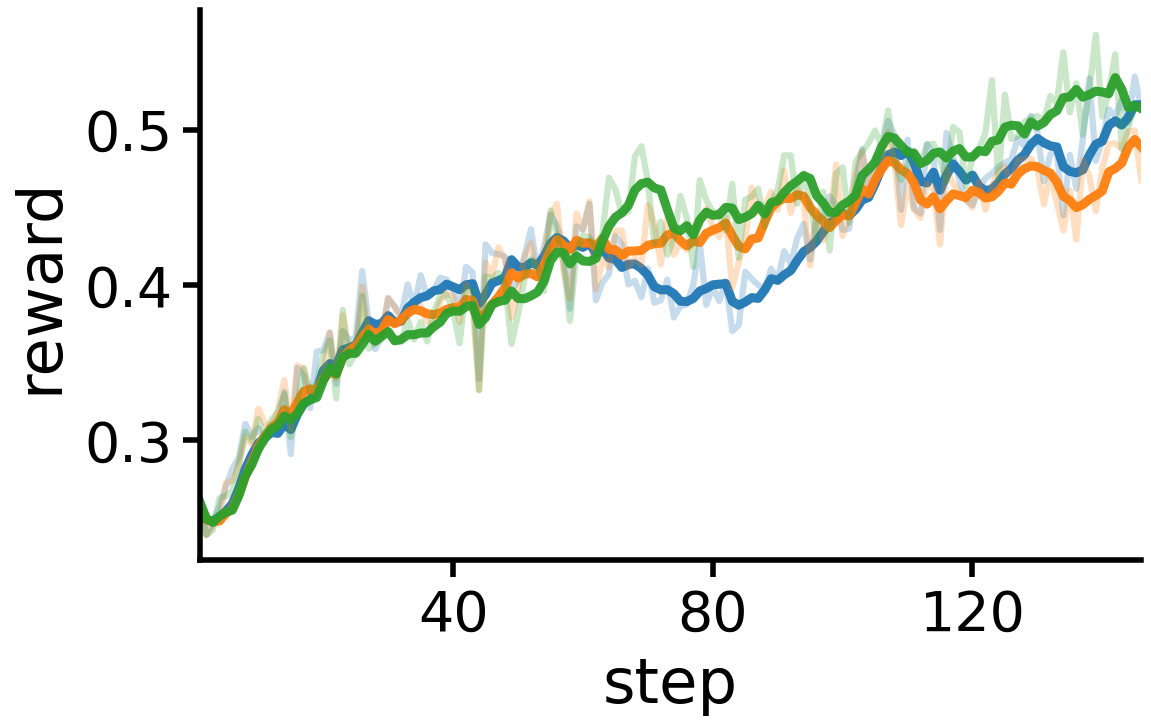} \\
\caption{Observation experiments on Qwen2.5-Math-7B trained on DAPO-17k. Starting from a rollout batch size of $512\times 8$, we additionally mix in 64 temperature-scaled samples per update. The left column varies the global sampling temperature in vanilla GRPO without separating positive and negative samples, while the middle and right columns isolate their effects under high- and low-temperature regimes. Changing temperature alone does not prevent entropy collapse; the entropy effect depends critically on both sample sign and temperature.}
\label{fig:obs_temp_sign}
\end{figure}

To better understand how different sampled trajectories affect entropy during RL post-training, we conduct controlled observation experiments by varying the sampling temperature and by separately constructing auxiliary updates from positive or negative samples. The results are shown in Fig.~\ref{fig:obs_temp_sign}. Three observations emerge.

\textbf{Increasing temperature does not prevent entropy collapse.}
We vary the global sampling temperature in vanilla GRPO without decoupling positive and negative samples. As shown in Fig.~\ref{fig:obs_temp_sign}, different temperatures slightly alter the entropy trajectory, but entropy still decreases monotonically under both high- and low-temperature sampling. This shows that temperature alone is insufficient to stabilize exploration.

\textbf{High-temperature positive and negative samples induce opposite entropy dynamics.}
Under high-temperature sampling, positive samples increase entropy while maintaining competitive reward, whereas negative samples accelerate entropy collapse. As shown in the middle column of Fig.~\ref{fig:obs_temp_sign}, high-temperature positive samples produce a sustained entropy rise, while high-temperature negative samples drive entropy down even faster than GRPO, indicating that they reinforce rather than mitigate collapse.

\textbf{The trend reverses under low-temperature sampling.}
Under low-temperature sampling, the entropy effect reverses: negative samples become entropy-increasing, whereas positive samples make entropy collapse more severe. However, although low-temperature negative samples can raise entropy, they also cause rapid reward collapse, making them unsuitable for stable training.

Taken together, these results reveal a temperature--sign asymmetry in sample-level entropy dynamics: the entropy effect is jointly determined by sample sign and temperature. Moreover, the failure of temperature-only control in vanilla GRPO shows that entropy regulation does not arise from temperature alone, but from its interaction with positive and negative samples. Motivated by this asymmetry, we next study a simplified bandit setting to characterize how temperature perturbs entropy drift and why high-temperature positive samples can counteract entropy collapse.

\begin{assumption} 
The actor policy $\pi_\theta$ is a tabular softmax policy; the reward is binary with $R(s,a)=1$ for positive samples; the logit gradients are orthogonal; and the update step is sufficiently small so that the one-step entropy variation admits a first-order approximation. The policy
is updated via
\begin{equation}
\theta_{k+1} = \theta_k + \eta \mathbb{E}_{a\sim\pi_{\theta_k}^{T}(\cdot|s)}
\left[R(s, a)\cdot \nabla_\theta \log \pi_{\theta_k}(a|s)\right].
\end{equation}
\end{assumption}

\begin{theorem}
    \label{theo:entropy}
    Denote the function
\begin{equation}
\Delta \mathcal{H}_{\theta_k}(T) = -\eta \cdot \sum_{a^*\in A^*}\pi_{\theta_k}^{T}(a^*|s)\times
\mathrm{Cov}_{\pi_{\theta_k}}\!\left(z_k(s,a'), \delta_{a',a^*}-\pi_{\theta_k}(a'|s)\right).
\end{equation}
Then we have
\begin{equation}
\Delta \mathcal{H}_{\theta_k}(T)\approx \mathcal{H}(\pi_{\theta_{k+1}}) - \mathcal{H}(\pi_{\theta_{k}}),
\end{equation}
and
\begin{equation}
\boxed{\frac{\textbf{d}\Delta \mathcal{H}_{\theta_k}(T)}{\textbf{d}T}\vert_{T=1} > 0} \quad \text{if} \quad \Delta \mathcal{H}_{\theta_k}(T)\vert_{T=1} < 0.
\end{equation}
\end{theorem}

The proof is provided in Appendix \ref{app:key theorem}. Theorem~\ref{theo:entropy} provides a theoretical explanation for the asymmetry observed above. Specifically, when the entropy drift becomes negative, increasing the sampling temperature under positive-sample updates introduces a positive correction to the entropy dynamics, thereby counteracting entropy collapse. Motivated by this result, we propose \textbf{SCOPE-RL}, which explicitly leverages temperature-adjusted positive samples as a lightweight regularization signal to stabilize entropy and sustain exploration during RL post-training.

\subsection{SCOPE-RL}

\begin{wrapfigure}{r}{0.45\linewidth}
	\centering
	\vspace{-0.8\baselineskip}
	\includegraphics[width=\linewidth]{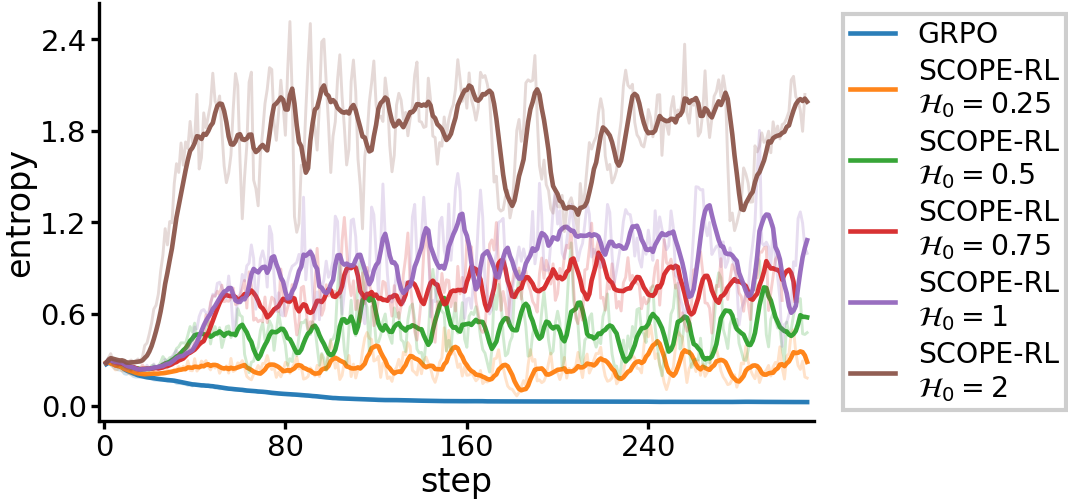}
	\vspace{-1\baselineskip}
	\caption{SCOPE-RL can not only alleviate entropy collapse, but also stably control entropy at different levels}
	\label{fig:main_entropy}
\end{wrapfigure}
Building on the above analysis, \textbf{SCOPE-RL} achieves entropy control by introducing a small auxiliary policy-gradient term constructed from temperature-adjusted positive samples. The core idea is straightforward but effective: when the policy entropy $\mathcal{H}(\pi_{\theta_{\rm old}})$ of the previous step falls below a target threshold $\mathcal{H}_0$, SCOPE-RL draws a small number of high-temperature positive samples to promote exploration; when the entropy exceeds $\mathcal{H}_0$, it instead uses a small number of low-temperature positive samples to maintain training stability. As shown in Fig. \ref{fig:main_entropy}, SCOPE-RL achieves entropy control by introducing a small auxiliary regularization term constructed from temperature-adjusted positive samples.

Formally, the SCOPE-RL objective is defined as:
\begin{equation}
\begin{split}
\small
\mathcal{J}_{\rm SCOPE}(\theta)=&
\mathcal{J}_{\rm GRPO}(\theta)+
\alpha\,
\mathbb{E}_{q\sim P(Q),\,\{o_i\}_{i=1}^{G'}\sim \pi_{\theta_{\rm old}}^{T}(O\mid q)} \\
&\left[
\frac{1}{G'}\sum_{i=1}^{G'}
\mathbf{1}[R(q,o_i)=1]\,
\frac{1}{|o_i|}\sum_{t=1}^{|o_i|}
\min\!\Big(
r_{i,t}(\theta),\,
\mathrm{clip}\!\big(r_{i,t}(\theta),1-\epsilon,1+\epsilon\big)
\Big)
\right].
\end{split}
\end{equation}
where
\begin{equation}
r_{i,t}(\theta)=
\frac{\pi_{\theta}(o_{i,t}\mid q,o_{i,<t})}
{\pi_{\theta_{\rm old}}(o_{i,t}\mid q,o_{i,<t})}, \quad \text{and} \quad T= \clip(1+\mathcal{H}_0-\mathcal{H}(\pi_{\theta_{\rm old}}), 0.8, 1.2).
\end{equation}

SCOPE-RL contains two key components: 
\begin{itemize}
    \item \emph{Positive samples as regularization}: motivated by the observations and analysis above, SCOPE-RL retains only positive samples in the auxiliary branch, as they provide a more stable signal for entropy control under temperature adjustment.
    \item \emph{Temperature as regularization}: when entropy falls below the target threshold, high-temperature samples are used to encourage exploration; when entropy rises above the threshold, low-temperature samples are used to restore stability.
\end{itemize}

\begin{table*}[t]
 \caption{Main results on mathematical reasoning benchmarks. SCOPE-RL consistently improves performance across different base models.}
 \small
 \label{tab:main_results_math}
 \centering
 \resizebox{\textwidth}{!}{%
  \renewcommand{\arraystretch}{1.4}
  \begin{tabular}{
    >{\raggedright\arraybackslash}m{3.7cm}
    *{7}{>{\centering\arraybackslash}m{1.6cm}} |
    >{\centering\arraybackslash}m{1.6cm}
   }
   \toprule
   \textbf{Methods} & \textbf{AIME24} & \textbf{AIME25} & \textbf{AMC23} & \textbf{GSM8K} & \textbf{MATH} & \textbf{Minerva} & \textbf{Olympiad} & \textbf{Average} \\
   \midrule
    \rowcolor{gray!10} Qwen2.5-Math-7B & 15.5 & 7.81 & 42.1 & 65.4 & 59.4 & 11.0 & 26.7 & 32.56 \\
    +GRPO  & 32.1 & 11.0 & 72.4 & 88.7 & 80.6 & 34.6 & 41.8 & 51.60 \\
    +Entropy-Reg $(\lambda=0.01)$ & 34.2 & 10.1 & 73.1 & 88.7 & 79.8 & 35.3 & 40.8 & 51.71 \\
    +Entropy-Reg $(\lambda=0.03)$ & 31.4 & 10.1 & 74.3 & 87.0 & 78.8 & 35.7 & 40.4 & 51.10 \\
    +Entropy-Reg $(\lambda=0.05)$ & 21.6 & 10.8 & 63.6 & 86.5 & 77.0 & 31.9 & 39.2 & 47.23 \\
    +DAPO & 33.3 & 13.0 & 73.9 & 87.9 & 79.6 & 34.6 & 40.8 & 51.87 \\
    +Clip higher $(\epsilon_{\text{high}}=0.4)$ & 29.8 & 9.58 & 64.1 & 80.6 & 76.2 & 30.1 & 40.4 & 47.25 \\
    +Clip higher $(\epsilon_{\text{high}}=0.5)$ & 24.0 & 7.92 & 61.8 & 84.9 & 76.4 & 32.7 & 38.9 & 46.66 \\
    +Entropy-Adv & 31.3 & 11.4 & 72.1 & 87.8 & 78.8 & 37.5 & 42.1 & 51.57 \\
    +CISPO & 35.0 & 13.8 & 75.1 & 89.7 & 80.2 & 37.1 & 41.6 & 53.21 \\
    +KL-cov & 32.1 & 13.5 & 73.9 & \textbf{89.9} & 79.8 & 35.6 & 41.7 & 52.36 \\
    \rowcolor{blue!5}
    +SCOPE-RL & \textbf{36.4} & \textbf{18.9} & \textbf{75.3} & 89.5 & \textbf{82.6} & \textbf{38.2} & \textbf{43.0} & \textbf{54.84} \\
    \noalign{\vspace{-1.5mm}}
    \rowcolor{blue!5}
    \qquad $\Delta$ vs. GRPO
    & \textcolor{green!70!black}{\textbf{(+4.3)}}
    & \textcolor{green!70!black}{\textbf{(+7.9)}}
    & \textcolor{green!70!black}{\textbf{(+2.9)}}
    & \textcolor{green!70!black}{\textbf{(+0.8)}}
    & \textcolor{green!70!black}{\textbf{(+2.0)}}
    & \textcolor{green!70!black}{\textbf{(+3.6)}}
    & \textcolor{green!70!black}{\textbf{(+1.2)}}
    & \textcolor{green!70!black}{\textbf{(+3.23)}} \\
   \midrule
    \rowcolor{gray!10} Qwen2.5-7B & 7.91 & 5.31 & 36.2 & 88.5 & 64.4 & 22.0 & 29.3 & 36.23 \\
    +GRPO  & 17.1 & 7.60 & 65.8 & 92.3 & 75.6 & 36.8 & 38.8 & 47.71 \\
    +DAPO & \textbf{17.7} & 9.06 & 68.0 & 92.1 & 76.4 & 37.1 & 40.1 & 48.64 \\
    +Entropy-Reg $(\lambda=0.01)$ & 13.6 & 8.85 & 67.4 & 92.3 & 76.8 & 35.5 & 39.1 & 47.65 \\
    +Entropy-Adv & 14.8 & 10.1 & 67.3 & 91.9 & 76.6 & \textbf{38.2} & 37.5 & 48.06 \\
    +CISPO & 15.1 & 8.10 & 66.6 & 92.3 & 77.6 & 37.1 & 38.2 & 47.86 \\
    +KL-cov & 16.9 & 11.3 & 65.5 & 91.5 & 76.8 & \textbf{38.2} & 39.1 & 48.47 \\
    \rowcolor{blue!5}
    +SCOPE-RL & 17.5 & \textbf{12.9} & \textbf{69.3} & \textbf{92.9} & \textbf{78.0} & 37.8 & \textbf{40.3} & \textbf{49.81} \\
    \noalign{\vspace{-1.5mm}}
    \rowcolor{blue!5}
    \qquad $\Delta$ vs. GRPO
    & \textcolor{green!70!black}{\textbf{(+0.4)}}
    & \textcolor{green!70!black}{\textbf{(+5.3)}}
    & \textcolor{green!70!black}{\textbf{(+3.5)}}
    & \textcolor{green!70!black}{\textbf{(+0.6)}}
    & \textcolor{green!70!black}{\textbf{(+2.4)}}
    & \textcolor{green!70!black}{\textbf{(+1.0)}}
    & \textcolor{green!70!black}{\textbf{(+1.5)}}
    & \textcolor{green!70!black}{\textbf{(+2.10)}} \\
   \midrule
    \rowcolor{gray!10} Qwen3-4B & 36.4 & 22.7 & 71.9 & 93.9 & 84.8 & 42.3 & 47.2 & 57.03 \\
    +GRPO  & 52.9 & 41.5 & 86.1 & \textbf{95.2} & 92.0 & 46.7 & 60.0 & 67.77 \\
    +DAPO & 53.1 & 41.7 & 87.3 & 94.5 & 92.0 & 47.1 & 61.1 & 68.04 \\
    +Entropy-Reg $(\lambda=0.01)$ & 52.4 & 42.6 & 88.1 & 94.6 & 91.6 & 46.3 & 60.1 & 67.96 \\
    +Entropy-Adv & 51.6 & 41.7 & 87.0 & 94.2 & 91.2 & 46.0 & 58.4 & 67.16 \\
    +CISPO & 52.7 & 42.6 & 89.3 & 94.7 & 91.4 & 47.8 & 60.3 & 68.40 \\
    +KL-cov & 53.9 & 42.3 & 88.7 & 94.8 & 92.4 & 46.3 & 60.0 & 68.34 \\
    \rowcolor{blue!5}
    +SCOPE-RL & \textbf{56.5} & \textbf{43.7} & \textbf{89.7} & 95.0 & \textbf{92.8} & \textbf{48.9} & \textbf{61.5} & \textbf{69.73} \\
    \noalign{\vspace{-1.5mm}}
    \rowcolor{blue!5}
    \qquad $\Delta$ vs. GRPO
    & \textcolor{green!70!black}{\textbf{(+3.4)}}
    & \textcolor{green!70!black}{\textbf{(+2.2)}}
    & \textcolor{green!70!black}{\textbf{(+3.6)}}
    & \textcolor{red!70!black}{\textbf{(-0.2)}}
    & \textcolor{green!70!black}{\textbf{(+0.8)}}
    & \textcolor{green!70!black}{\textbf{(+2.2)}}
    & \textcolor{green!70!black}{\textbf{(+1.5)}}
    & \textcolor{green!70!black}{\textbf{(+1.96)}} \\
   \bottomrule
  \end{tabular}
 }
\end{table*}

\section{Experiments}
To validate SCOPE-RL, we implement it within the EasyR1 and VeRL frameworks \cite{zheng2025easyr1,sheng2025hybridflow} and compare it against representative RL baselines, including GRPO~\cite{shao2024deepseekmath}, entropy-based methods such as Entropy-Reg~\cite{hou2025advancing}, Entropy-Adv~\cite{ cheng2025reasoning}, and KL-cov~\cite{cui2025entropy}, as well as clipping- or entropy-related approaches including DAPO~\cite{yu2025dapo} and CISPO~\cite{ chen2025minimax}. Experiments are conducted on Qwen2.5-7B, Qwen2.5-Math-7B, and Qwen3-4B \cite{yang2024qwen2, yang2025qwen3}, using DAPO-17K \cite{yu2025dapo} for training. We evaluate both in-domain mathematical reasoning and out-of-domain generalization on seven math benchmarks, including AIME24, AIME25 \cite{hf_aime2024}, AMC23 \cite{lightman2023lets}, GSM8K \cite{cobbe2021gsm8k}, MATH \cite{lightman2023lets}, Minerva Math \cite{lewkowycz2022solving}, and Olympiad \cite{lightman2023lets}, as well as three knowledge-intensive benchmarks, ARC-Challenge~\cite{clark2018think}, MMLU-Pro\cite{wang2024mmlu}, and SuperGPQA~\cite{du2025supergpqa}. The detailed implementation is shown in Appendix \ref{app:detail}.

\begin{table*}[t]
 \caption{Generalization results on knowledge-intensive benchmarks, including ARC-Challenge, MMLU-Pro, and SuperGPQA, where SCOPE-RL demonstrates leading performance across RL methods.}
 \label{tab:main_results_knowledge}
 \centering
 \resizebox{\textwidth}{!}{%
  \renewcommand{\arraystretch}{1.35}
  \begin{tabular}{
    >{\raggedright\arraybackslash}m{2.4cm}
    *{3}{>{\centering\arraybackslash}m{1.3cm}}
    *{3}{>{\centering\arraybackslash}m{1.3cm}}
    *{3}{>{\centering\arraybackslash}m{1.3cm}}
   }
   \toprule
   \multirow{2}{*}{\textbf{Methods}} 
   & \multicolumn{3}{c}{\textbf{Qwen2.5-Math-7B}}
   & \multicolumn{3}{c}{\textbf{Qwen2.5-7B}}
   & \multicolumn{3}{c}{\textbf{Qwen3-4B}} \\
   \cline{2-10}
   & ARC & MMLU & S-GPQA
   & ARC & MMLU & S-GPQA
   & ARC & MMLU & S-GPQA \\
   \midrule
   Base & 69.9 & 34.3 & 19.4 & 86.3 & 50.0 & 18.6 & 92.9 & 72.6 & 32.8 \\
   +GRPO & 78.9 & 44.8 & 25.8 & 89.3 & 58.2 & 27.0 & \textbf{94.6} & 74.1 & 38.4 \\
   +DAPO & 80.9 & \textbf{48.6} & 26.0 & 90.9 & 59.8 & 24.8 & 92.6 & 75.0 & 37.0 \\
   +Entropy-Reg & 79.6 & 42.7 & 26.2 & 89.6 & 58.2 & 26.8 & 93.9 & 73.7 & 38.2 \\
   +Entropy-Adv & 80.3 & 47.3 & 26.2 & 90.6 & 57.7 & 26.6 & 93.6 & 74.8 & 37.8 \\
   +CISPO & 80.9 & 44.1 & 26.2 & 89.9 & 57.8 & 25.8 & 92.3 & 75.1 & 36.8 \\
   +KL-cov & 79.9 & 48.0 & 26.6 & 90.9 & 59.1 & 26.0 & 93.3 & 75.5 & 37.6 \\
   \rowcolor{blue!5}
   +SCOPE-RL & \textbf{80.9} & 48.0 & \textbf{26.8} & \textbf{91.6} & \textbf{60.7} & \textbf{27.2} & 94.3 & \textbf{76.1} & \textbf{38.6} \\
   \bottomrule
  \end{tabular}
 }
\end{table*}
\begin{table*}
    \caption{Comparison of Pass@$k$ performance between the base model, GRPO, and SCOPE-RL across AIME24 and AIME-25. The accompanying figures show the evolution of Pass@1024 of SCOPE-RL and GRPO on AIME24 and AIME25 during training.}
    \label{tab:passk}
    \centering
    \resizebox{\textwidth}{!}{
        \renewcommand{\arraystretch}{1.4}
        \begin{tabular}{
        >{\raggedright\arraybackslash}m{2.8cm}  
        *{4}{>{\centering\arraybackslash}m{3.2cm}}
        }
        \toprule
        \multirow{2}{*}{\textbf{Benchmarks}} & \multicolumn{2}{c}{\textbf{AIME24}} & \multicolumn{2}{c}{\textbf{AIME25}} \\
        \cmidrule{2-5}
        & Pass@512 & Pass@1024 & Pass@512 & Pass@1024\\
        \midrule
        \rowcolor{gray!10} Qwen2.5-7B & \textbf{71.7} & \underline{80.0} & \underline{61.7} & \underline{73.3}\\
        \quad +GRPO  & 63.3 & 73.3 & 60.0 & 63.3\\
        \rowcolor{blue!5} \quad +SCOPE-RL & \underline{70.0} & \textbf{86.7} & \textbf{66.7} & \textbf{76.7}\\
        \bottomrule
        \end{tabular}
    }\\
    \includegraphics[width=0.45\linewidth]{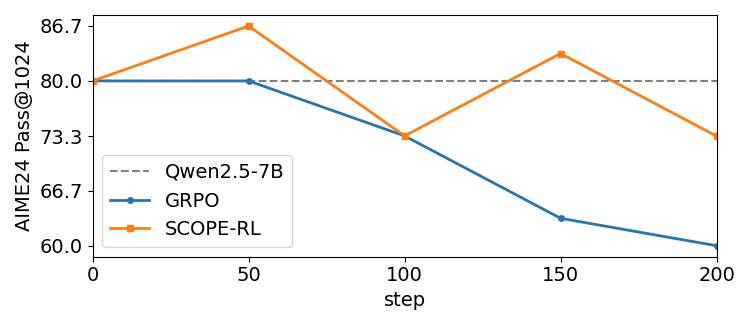}
    \includegraphics[width=0.45\linewidth]{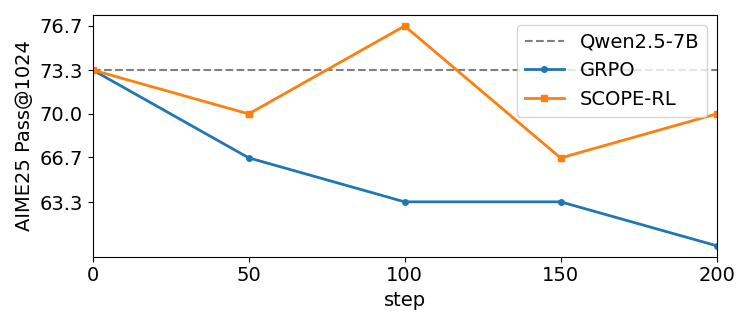}
    \vspace{-10pt}
\end{table*}

\subsection{Main Results}

\textbf{In-domain Mathematical Reasoning.}
Table~\ref{tab:main_results_math} shows that SCOPE-RL achieves the best average performance on all three base models, improving over GRPO from 51.60 to 54.84 on Qwen2.5-Math-7B, from 47.71 to 49.81 on Qwen2.5-7B, and from 67.77 to 69.73 on Qwen3-4B. It is particularly strong on challenging benchmarks such as AIME24, AIME25, and MATH, and achieves the strongest overall compared with other RL baselines. 

\textbf{Out-of-domain Generalization.}
Table~\ref{tab:main_results_knowledge} shows that SCOPE-RL also transfers well beyond math training data. On Qwen2.5-7B, it achieves the best results on all three knowledge benchmarks, reaching 91.6 on ARC-Challenge, 60.7 on MMLU-Pro, and 27.2 on SuperGPQA. This suggests that the benefit of entropy control is not limited to in-domain mathematical reasoning.

\textbf{Pass@$k$ Performance and Reasoning Capability.}
Table~\ref{tab:passk} provides a complementary view from Pass@$k$ evaluation. On Qwen2.5-7B, GRPO underperforms the base model at Pass@1024 on both AIME24 and AIME25, while SCOPE-RL improves them to 86.7 and 76.7, respectively. The accompanying curves show the same trend dynamically: as training proceeds, Pass@1024 under GRPO degrades steadily, while SCOPE-RL maintains clearly stronger many-sample performance throughout training. This is consistent with our claim that stable entropy control preserves useful exploration, which is particularly important for the reasoning capability.

\textbf{Quantitative Analysis of Exploration Degree.}
Table~\ref{tab:entropy} suggests a non-monotonic relationship between entropy level and performance in our setting. When varying the entropy target, SCOPE-RL achieves the best average performance at $\mathcal{H}_0=0.50$ with 54.84, outperforming $\mathcal{H}_0=0.25$ (53.00) and $\mathcal{H}_0=0.75$ (53.45). These results indicate that exploration is most beneficial when maintained within an appropriate range.

\begin{figure*}[t]
	\centering
	\includegraphics[width=0.9\linewidth]{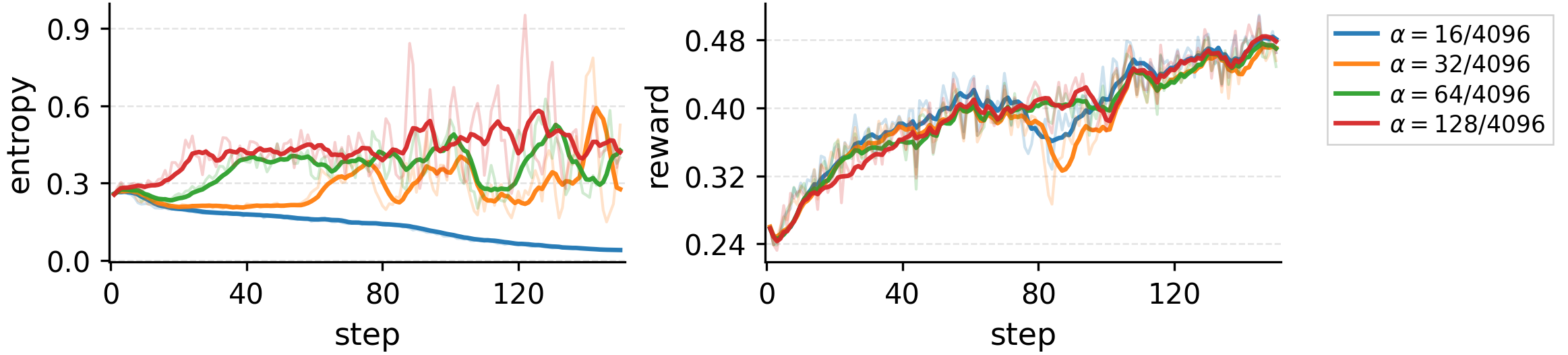}
	\caption{Entropy and reward curves of SCOPE-RL under different auxiliary sample ratios $\alpha$ on Qwen2.5-Math-7B, trained on DAPO-17K with a rollout batch size of $512 \times 8$ and $\mathcal{H}_0=0.5$.}
	\label{fig:alpha_study}
\end{figure*}

\textbf{Effectiveness and Efficiency.} An important consideration of SCOPE-RL is its effectiveness and efficiency trade-off. To study this, we vary the auxiliary sample ratio $\alpha$, which in our implementation also controls the effective regularization strength. As shown in Fig.~\ref{fig:alpha_study}, compared with GRPO, whose entropy steadily collapses, SCOPE-RL consistently raises the entropy level and slows down collapse even with a very small auxiliary budget, indicating only marginal additional sampling cost. Meanwhile, a larger $\alpha$ does not necessarily improve optimization: although $\alpha=128/4096$ yields higher entropy, it also introduces substantially larger fluctuations, suggesting reduced training stability. In contrast, moderate values such as $\alpha=32/4096$ and $\alpha=64/4096$ are already sufficient to prevent entropy collapse while maintaining rewards comparable to GRPO, using less than $2\%$ additional auxiliary samples. In practice, we adopt $\alpha=64/4096=1/64$ as the default setting, which provides effective entropy control with limited overhead.

\textbf{Additional Results.}
Appendix~\ref{app:more_results} provides further evidence beyond the main experiments. Three-seed experiments on Llama-3.1-8B-Instruct under matched rollout budgets examine robustness across model families and training runs (Appendix~\ref{app:llama_results}); expanded Pass@32 comparisons cover four base models and several anti-collapse baselines (Appendix~\ref{app:pass32}). We also report prompt-level correctness dynamics to complement the entropy analysis (Appendix~\ref{app:useful_exploration}) and test entropy control under different clipping settings (Appendix~\ref{app:clip_robustness}).

\begin{table*}[t]
\centering
\caption{SCOPE-RL under different entropy targets $\mathcal{H}_0$, which correspond to different degrees of exploration. The overall performance is maximized at a moderate entropy level.}
\label{tab:entropy}
 \resizebox{\textwidth}{!}{%
  \renewcommand{\arraystretch}{1.4}
  \begin{tabular}{
    >{\raggedright\arraybackslash}m{4.0cm}  
    *{7}{>{\centering\arraybackslash}m{1.6cm}}| 
    >{\centering\arraybackslash}m{1.6cm}    
   }
   \toprule
   \textbf{Benchmarks} & \textbf{AIME24} & \textbf{AIME25} & \textbf{AMC}  & \textbf{GSM8K} & \textbf{MATH} & \textbf{Minerva} & \textbf{Olympiad} & \textbf{Average} \\
   \midrule
   \rowcolor{gray!10} Qwen2.5-math-7B  & 15.5 & 7.81 & 42.1 & 65.4 & 59.4 & 11.0 & 26.7 & 32.56 \\
   \rowcolor{green!5} \quad +SCOPE-RL $\mathcal{H}_0=0.25$  & \textbf{37.9} & 11.3 & 74.4 & 89.4 & \underline{79.8} & 36.0 & 42.2 & 53.00 \\
   \rowcolor{green!5} \quad +SCOPE-RL $\mathcal{H}_0=0.50$  & \underline{36.4} & \textbf{18.9} & \underline{75.3} & \textbf{89.5} & \textbf{82.6} & \textbf{38.2} & \textbf{43.0} & \textbf{54.84} \\
   \rowcolor{green!5} \quad +SCOPE-RL $\mathcal{H}_0=0.75$ & 33.6 & \underline{15.1} & \textbf{77.0} & \underline{89.4} & 79.2 & 37.5 & \underline{42.4} & \underline{53.45} \\
   \rowcolor{green!5} \quad +SCOPE-RL $\mathcal{H}_0=1.00$  & 33.2 & \textbf{15.6} & 74.8 & 88.7 & 79.6 & \underline{37.9} & 42.1 & 53.13 \\
   \bottomrule
  \end{tabular}
 }
 
\end{table*}

\begin{table*}[t]
\centering
\caption{Component ablations of SCOPE-RL on Qwen2.5-Math-7B. Removing either temperature-adjusted or positive sampling  degrades performance and entropy control.}
\label{tab:ablation}
\resizebox{\textwidth}{!}{
\renewcommand{\arraystretch}{1.35}
\begin{tabular}{
>{\raggedright\arraybackslash}m{3.0cm}
*{7}{>{\centering\arraybackslash}m{1.6cm}}
>{\centering\arraybackslash}m{2.6cm}
}
\toprule
\textbf{Setting} & \textbf{AIME24} & \textbf{AIME25} & \textbf{AMC} & \textbf{GSM8K} & \textbf{MATH} & \textbf{Minerva} & \textbf{Olympiad} & \textbf{Average} \\
\textbf{Entropy step} & \textbf{50} & \textbf{100} & \textbf{150} & \textbf{200} & \textbf{250} & \textbf{300} & \textbf{350} & \textbf{Entropy trends} \\
\midrule
\rowcolor{red!5} 
& 31.4 & 11.8 & 74.5 & 88.9 & 79.0 & 33.5 & 40.1 & 51.31 {\color{red}(-3.53)} \\
\rowcolor{red!5} 
\multirow{-2}{*}{w/o temperature}& 0.080 & 0.036 & 0.026 & 0.024 & 0.022 & 0.019 & 0.018 & Entropy collapse \\
\rowcolor{red!5} 
& 32.7 & 11.0 & 73.2 & 88.1 & 79.0 & 35.7 & 41.2 & 51.56 {\color{red}(-3.28)} \\
\rowcolor{red!5} 
\multirow{-2}{*}{positive+negative}& 0.114 & 0.041 & 0.027 & 0.024 & 0.024 & 0.022 & 0.020 & Entropy collapse \\
\rowcolor{red!5} 
& 28.9 & 12.7 & 65.1 & 89.5 & 77.6 & 33.8 & 40.4 & 49.71 {\color{red}(-5.12)} \\
\rowcolor{red!5} 
\multirow{-2}{*}{only negative} & 0.492 & 1.176 & 5.224 & 5.948 & 7.235 & - & - & Entropy instability \\
\rowcolor{red!5} 
only regularizer
& - & - & - & - & - & - & - &  Training failed \\
\bottomrule
\end{tabular}
}
\end{table*}

\begin{figure}[t]
    \centering
    \includegraphics[width=0.9\linewidth]{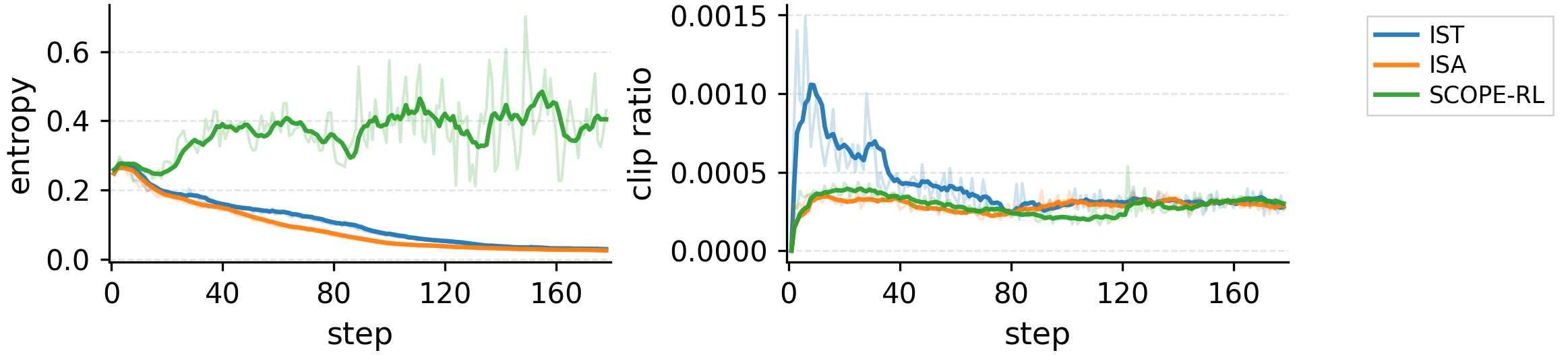}
    \caption{Importance-sampling ablations on entropy control and clipping behavior.}
    \label{fig:is_ablation}
\end{figure}

\section{Ablations}

We conduct ablation studies to understand the contribution of each design choice in SCOPE-RL and investigate the underlying mechanism behind its entropy control. Specifically, we analyze the necessity of temperature adaptation and positive-sample filtering, as well as the role of the intentional sampling-distribution mismatch through importance-sampling ablations.

\subsection{Component Ablations}
We ablate the key components of SCOPE-RL in Table~\ref{tab:ablation}. \textbf{Temperature adjustment.} Removing temperature-adjusted sampling reduces the average score from 54.84 to 51.31 and leads to entropy collapse, showing that the positive-only auxiliary branch is insufficient by itself. \textbf{Positive-sample filtering.} Including both positive and negative auxiliary samples reduces the average score to 51.56 and again leads to entropy collapse. Using only negative auxiliary samples produces unstable entropy and weaker performance, while using only the auxiliary regularizer fails to train. Together, these results support the combination of temperature adaptation and positive-sample filtering used in SCOPE-RL.
\subsection{Importance-Sampling Ablations}
The auxiliary samples of SCOPE-RL are drawn from $\pi_{\theta_{\rm old}}^{T}$, but the likelihood ratio in SCOPE-RL uses $\pi_{\theta_{\rm old}}$. \textbf{This mismatch is intentional: temperature changes which positive trajectories enter the auxiliary update, and correcting for the sampling distribution may remove the reweighting that supports entropy control.} We therefore test whether this mismatch matters empirically, while separating its effect from a possible change in clipping behavior. We consider two importance-sampling variants of the auxiliary objective: importance sampling on the likelihood ratio (IST) and importance sampling on the advantage (ISA):
\begin{equation}
\begin{aligned}
r_{i,t}(\theta)
&\leftarrow
\frac{
\pi_{\theta}(o_{i,t}\mid q,o_{i,<t})
}{
\pi_{\theta_{\rm old}}^{T}(o_{i,t}\mid q,o_{i,<t})
},
&& \text{(IST)}
\\[6pt]
\mathbf{1}[R(q,o_i)=1]
&\leftarrow
\frac{
\pi_{\theta_{\rm old}}(o_{i,t}\mid q,o_{i,<t})
}{
\pi_{\theta_{\rm old}}^{T}(o_{i,t}\mid q,o_{i,<t})
}
\mathbf{1}[R(q,o_i)=1].
&& \text{(ISA)}
\end{aligned}
\label{eq:importance_ablations}
\end{equation}
The first variant clips the corrected ratio. The second retains SCOPE-RL's original ratio and clipping rule, applying the correction as a token-wise weight instead. All other components are unchanged. Fig.~\ref{fig:is_ablation} shows that entropy declines under both importance-sampling variants while SCOPE-RL maintains higher entropy, and the clipping fractions further indicate that this degradation is not caused by altered clipping behavior, as entropy still declines when the original clipping rule is retained. \textbf{These results show that SCOPE-RL's entropy-control ability depends on the bias introduced by temperature-adjusted sampling: correcting this bias through importance weighting causes entropy control to disappear. SCOPE-RL achieves better downstream performance by confining the biased auxiliary update to a small, controlled fraction of training, so that it regulates entropy without dominating reward optimization.}


\section{Related Work}
Reinforcement learning (RL) has become a core paradigm for post-training large language models (LLMs), especially for improving reasoning beyond supervised fine-tuning \cite{team2025kimi1_5, guo2025deepseek, openai2023gpt4, team2024gemini1_5, wei2023instructiongpt}. Among recent reasoning-oriented RL methods, Group Relative Policy Optimization (GRPO) has emerged as a widely adopted backbone due to its simplicity, scalability, and effectiveness \cite{shao2024deepseekmath, liu2024deepseek, guo2025deepseek}. However, recent evidence suggests that RL post-training often sharpens behaviors already latent in the base model rather than consistently expanding its reasoning frontier \cite{yue2025does}, motivating growing interest in exploration during RL post-training. Entropy has long been viewed as a key proxy for exploration, as it reflects policy uncertainty and output diversity and helps prevent premature convergence \cite{schulman2017equivalence, haarnoja2018soft, nachum2017bridging, sutton1999policy, williams1992simple}. Accordingly, recent work has revisited entropy regulation in RL for LLMs as a means to sustain exploration and improve optimization \cite{cui2025entropy, hou2025advancing, shen2025entropy, cheng2025reasoning}.

Recent studies have identified entropy collapse as a major failure mode of GRPO-based training, where policy entropy decreases monotonically, sampled outputs become increasingly homogeneous, and exploration vanishes prematurely \cite{yu2025dapo, li2025disco, zhang2025edge}. Existing attempts to mitigate this issue mainly fall into two directions. The first introduces entropy bonuses into rewards, advantages, or auxiliary objectives, such as Entropy Regularization (Entropy-Reg) \cite{hou2025advancing}, KL-Cov \cite{cui2025entropy}, AEnt \cite{shen2025entropy}, Entropy Advantage (Entropy-Adv) \cite{cheng2025reasoning} and so on. The second relies on clipping-based or clip-higher strategies, such as DAPO \cite{yu2025dapo}, Clip-Cov \cite{cui2025entropy}, CISPO \cite{chen2025minimax}, CE-GPPO \cite{su2025gppo} and so on. Although both directions can alleviate entropy collapse to some extent, they often lead to unstable training dynamics, with entropy oscillating between collapse and explosion rather than remaining in a beneficial regime. In contrast, SCOPE does not directly manipulate entropy through bonuses or clipping heuristics. Instead, it exploits sample-level entropy dynamics, using a very small fraction of temperature-adaptive positive samples to achieve stable entropy control and enable quantitative analysis of entropy’s effect on training.

\section{Limitations}
Our theoretical analysis is developed under several simplifying assumptions, including a contextual-bandit abstraction, a tabular softmax policy, binary rewards, orthogonal logit gradients, and sufficiently small updates; it also analyzes the temperature derivative only locally at $T=1$. These assumptions are stronger than the full autoregressive setting of language modeling, but are adopted to isolate the local mechanism behind entropy control rather than to characterize all RL post-training dynamics. In particular, Theorem~\ref{theo:entropy} focuses on the regime $\Delta\mathcal{H}_{\theta_k}(1)<0$, which corresponds to the practically important setting where entropy decreases during a policy update and entropy collapse emerges. Under explicit entropy control, the opposite case $\Delta\mathcal{H}_{\theta_k}(1)>0$ may also arise; in this regime, the key inequality in Eq.~\eqref{eq:limit} no longer holds strictly, making the current proof inapplicable and a useful quantitative bound harder to derive. We therefore do not claim that the theory fully covers every controlled-entropy regime or establishes stability at arbitrary temperatures. Nevertheless, it provides a useful perspective on how temperature-adjusted positive samples reshape entropy dynamics and counteract entropy collapse. Beyond the theorem's scope, the empirical results in \textit{Observation and Analysis} (Fig.~\ref{fig:obs_temp_sign}) and \textit{Effectiveness and Efficiency} (Fig.~\ref{fig:alpha_study}) show effective entropy control in the evaluated training settings.

\section{Conclusion}
In this work, we investigate entropy collapse in RL post-training for reasoning LLMs, and our results suggest that stable exploration is important for effective optimization. We identify a temperature--sign asymmetry in sample-level entropy dynamics and, based on this insight, propose SCOPE-RL, which regulates entropy through temperature-adaptive positive-sample regularization. Experiments show that SCOPE-RL consistently improves both Pass@1 and Pass@$k$ over RL baselines, while enabling quantitative analysis of the exploration--performance trade-off. Overall, our results suggest that maintaining a proper level of entropy is key to improving reasoning performance in RL post-training. A promising direction for future work is to move beyond global entropy control and develop state-dependent exploration mechanisms that adapt to problem difficulty and reasoning stage, together with a more complete theory of entropy dynamics in token-level RL.

\section*{Acknowledgments}
This work is supported by the Zhongguancun Academy (Grant No.s C20250203) and Natural Science Foundation of Tianjin (No.24JCQNJC02170). 

\bibliographystyle{unsrt}
\bibliography{reference}
\appendix

\section{More Results}
\label{app:more_results}

\subsection{Additional Results on Llama-3.1-8B-Instruct}
\label{app:llama_results}

To further evaluate the robustness of SCOPE-RL across different model families and training randomness, we conduct additional experiments on Llama-3.1-8B-Instruct with three independent random seeds. For a fair comparison, the rollout budgets of all baselines are strictly controlled to be identical. We report the mean and standard deviation across the three runs in Table~\ref{tab:llama_results}.

\begin{table*}[h]
\centering
\caption{
Additional results on Llama-3.1-8B-Instruct over three independent random
seeds. All methods use identical rollout budgets. Results are reported as
mean $\pm$ standard deviation.
}
\label{tab:llama_results}
\renewcommand{\arraystretch}{1.25}
\setlength{\tabcolsep}{3.5pt}

\resizebox{0.94\textwidth}{!}{%
\begin{tabular}{lcccccc}
\toprule
\multirow{2}{*}{\textbf{Method}}
& \textbf{AIME24}
& \textbf{AIME25}
& \textbf{AMC23}
& \textbf{ARC}
& \textbf{CMIMC25}
& \textbf{GSM8K} \\
&
\textbf{HMMT25}
& \textbf{MATH500}
& \textbf{Minerva}
& \textbf{MMLU-Pro}
& \textbf{Olympiad}
& \textbf{SuperGPQA} \\
\midrule

\multirow{2}{*}{\textbf{SCOPE-RL}}
& $\mathbf{11.25 \pm 1.00}$
& $\mathbf{1.53 \pm 0.33}$
& $\mathbf{28.62 \pm 1.63}$
& $84.95 \pm 0.00$
& $\mathbf{4.87 \pm 1.34}$
& $\mathbf{88.32 \pm 0.40}$ \\
&
$\mathbf{1.81 \pm 1.15}$
& $\mathbf{54.33 \pm 1.21}$
& $\mathbf{29.17 \pm 0.56}$
& $\mathbf{53.63 \pm 1.19}$
& $\mathbf{19.95 \pm 0.31}$
& $\mathbf{19.47 \pm 0.64}$ \\

\addlinespace[1mm]

\multirow{2}{*}{GRPO}
& $7.74 \pm 0.34$
& $0.24 \pm 0.16$
& $26.72 \pm 0.68$
& $83.72 \pm 1.27$
& $3.88 \pm 0.16$
& $87.09 \pm 0.32$ \\
&
$0.90 \pm 0.24$
& $52.93 \pm 1.86$
& $25.73 \pm 1.33$
& $50.83 \pm 1.76$
& $19.16 \pm 2.23$
& $17.07 \pm 0.46$ \\

\addlinespace[1mm]

\multirow{2}{*}{DAPO}
& $6.73 \pm 0.95$
& $1.04 \pm 0.21$
& $25.21 \pm 1.52$
& $81.60 \pm 1.74$
& $4.20 \pm 0.32$
& $87.77 \pm 0.50$ \\
&
$0.21 \pm 0.21$
& $52.20 \pm 2.03$
& $24.88 \pm 0.21$
& $51.49 \pm 1.52$
& $18.47 \pm 0.09$
& $18.27 \pm 1.21$ \\

\addlinespace[1mm]

\multirow{2}{*}{KL-Cov}
& $8.58 \pm 0.79$
& $0.45 \pm 0.26$
& $22.45 \pm 2.35$
& $\mathbf{85.28 \pm 1.21}$
& $3.23 \pm 0.56$
& $88.22 \pm 0.19$ \\
&
$0.31 \pm 0.32$
& $50.93 \pm 2.14$
& $24.63 \pm 0.64$
& $52.03 \pm 0.92$
& $18.82 \pm 0.39$
& $17.47 \pm 1.10$ \\

\addlinespace[1mm]

CISPO
& \multicolumn{6}{c}{Training failed} \\

\bottomrule
\end{tabular}
}
\end{table*}

These additional results provide further evidence that the effectiveness of SCOPE-RL transfers across model families. Compared with the main experiments on Qwen2.5-7B, Qwen2.5-Math-7B, and Qwen3-4B, the evaluation on Llama-3.1-8B-Instruct extends the empirical study to a different model family and a larger model scale. Moreover, the three independent training runs quantify the variability induced by training randomness, while controlling all methods under the same rollout budget.

\subsection{Expanded Pass@32 Evaluation}
\label{app:pass32}

We further expand the Pass@$k$ evaluation to four base models spanning four models, including Qwen2.5-Math-7B, Qwen2.5-7B, Qwen3-4B, and Llama-3.1-8B-Instruct. In addition to GRPO, we compare SCOPE-RL against representative entropy- or clipping-based anti-collapse baselines, including DAPO, KL-Cov, and CISPO. The results are shown in Table~\ref{tab:pass32_expanded}.

\begin{table*}[h]
\centering
\caption{
Expanded Pass@32 evaluation across four base models spanning three model families. SCOPE-RL achieves the highest average Pass@32 performance on all evaluated base models.
}
\label{tab:pass32_expanded}

\small
\renewcommand{\arraystretch}{1.18}
\setlength{\tabcolsep}{4.5pt}

\begin{tabular}{llcccccc}
\toprule
\textbf{Base Model}
& \textbf{Method}
& \textbf{AIME24}
& \textbf{AIME25}
& \textbf{AMC23}
& \textbf{CMIMC25}
& \textbf{HMMT25}
& \textbf{Average} \\
\midrule

\multirow{5}{*}{\makecell{Qwen2.5-\\Math-7B}}
& \cellcolor{blue!5}\textbf{SCOPE-RL}
& \cellcolor{blue!5}\textbf{60.00}
& \cellcolor{blue!5}\textbf{40.00}
& \cellcolor{blue!5}\textbf{87.50}
& \cellcolor{blue!5}\textbf{20.00}
& \cellcolor{blue!5}\textbf{13.30}
& \cellcolor{blue!5}\textbf{44.16} \\

& GRPO
& 53.30
& 23.30
& 82.50
& 15.00
& 10.00
& 36.82 \\

& DAPO
& 50.00
& 33.33
& 82.50
& \textbf{20.00}
& 10.00
& 39.17 \\

& KL-Cov
& 53.33
& 26.67
& 82.50
& \textbf{20.00}
& 10.00
& 38.50 \\

& CISPO
& 50.00
& 26.67
& 80.00
& 15.00
& 6.67
& 35.67 \\

\midrule

\multirow{5}{*}{Qwen2.5-7B}
& \cellcolor{blue!5}\textbf{SCOPE-RL}
& \cellcolor{blue!5}\textbf{40.00}
& \cellcolor{blue!5}\textbf{36.67}
& \cellcolor{blue!5}\textbf{90.00}
& \cellcolor{blue!5}\textbf{32.50}
& \cellcolor{blue!5}\textbf{13.33}
& \cellcolor{blue!5}\textbf{42.50} \\

& GRPO
& 26.67
& 30.00
& 87.50
& 25.00
& 10.00
& 35.83 \\

& DAPO
& 36.67
& 30.00
& 87.50
& 27.50
& \textbf{13.33}
& 39.00 \\

& KL-Cov
& 30.00
& 26.67
& 80.00
& 30.00
& \textbf{13.33}
& 36.00 \\

& CISPO
& \multicolumn{6}{c}{Training failed} \\

\midrule

\multirow{5}{*}{Qwen3-4B}
& \cellcolor{blue!5}\textbf{SCOPE-RL}
& \cellcolor{blue!5}\textbf{86.67}
& \cellcolor{blue!5}73.33
& \cellcolor{blue!5}\textbf{100.00}
& \cellcolor{blue!5}\textbf{60.00}
& \cellcolor{blue!5}\textbf{56.67}
& \cellcolor{blue!5}\textbf{75.33} \\

& GRPO
& 80.00
& 63.33
& 97.50
& 47.50
& 53.33
& 68.33 \\

& DAPO
& 76.67
& \textbf{76.67}
& 97.50
& 52.50
& 50.00
& 70.67 \\

& KL-Cov
& 83.33
& 66.67
& 97.50
& 50.00
& 53.33
& 70.17 \\

& CISPO
& 76.67
& 70.00
& 97.50
& 50.00
& 46.67
& 68.17 \\

\midrule

\multirow{5}{*}{\makecell{Llama-3.1-\\8B-Instruct}}
& \cellcolor{blue!5}\textbf{SCOPE-RL}
& \cellcolor{blue!5}\textbf{36.67}
& \cellcolor{blue!5}\textbf{20.00}
& \cellcolor{blue!5}\textbf{80.00}
& \cellcolor{blue!5}15.00
& \cellcolor{blue!5}\textbf{10.00}
& \cellcolor{blue!5}\textbf{32.33} \\

& GRPO
& 33.33
& 13.33
& 72.50
& 12.50
& 6.67
& 27.67 \\

& DAPO
& 33.33
& 13.33
& 75.00
& 15.00
& \textbf{10.00}
& 29.33 \\

& KL-Cov
& \textbf{36.67}
& 10.00
& 72.50
& \textbf{17.50}
& 6.67
& 28.67 \\

& CISPO
& \multicolumn{6}{c}{Training failed} \\

\bottomrule
\end{tabular}
\end{table*}

SCOPE-RL achieves the highest average Pass@32 performance on all four base models. Specifically, it reaches 44.16 on Qwen2.5-Math-7B, 42.50 on Qwen2.5-7B, 75.33 on Qwen3-4B, and 32.33 on Llama-3.1-8B-Instruct, consistently outperforming GRPO and the representative anti-collapse baselines.

\subsection{Whether Preserved Entropy Reflects Useful Reasoning Exploration}
\label{app:useful_exploration}

To examine whether the entropy preserved by SCOPE-RL corresponds to useful reasoning exploration, we analyze prompt-level training dynamics on Qwen2.5-Math-7B trained with DAPO-17K. For each epoch, prompts are categorized as \emph{all-correct}, \emph{mixed}, or \emph{all-wrong} according to the correctness of their sampled responses. The results are shown in Table~\ref{tab:prompt_exploration}.

\begin{table}[h]
\centering
\caption{
Prompt-level training dynamics of GRPO and SCOPE-RL on Qwen2.5-Math-7B over four epochs.
}
\label{tab:prompt_exploration}

\small
\renewcommand{\arraystretch}{1.15}
\setlength{\tabcolsep}{5pt}

\begin{tabular}{lccccc}
\toprule
\textbf{Method}
& \textbf{Epoch}
& \textbf{Entropy}
& \textbf{All-correct}
& \textbf{Mixed}
& \textbf{All-wrong} \\
\midrule

GRPO
& 1
& 0.192
& 0.105
& 0.484
& 0.411 \\

\rowcolor{blue!5}
\textbf{SCOPE-RL}
& 1
& \textbf{0.310}
& 0.093
& 0.484
& 0.423 \\

\midrule

GRPO
& 2
& 0.151
& 0.163
& 0.442
& 0.395 \\

\rowcolor{blue!5}
\textbf{SCOPE-RL}
& 2
& \textbf{0.418}
& 0.161
& \textbf{0.443}
& 0.396 \\

\midrule

GRPO
& 3
& 0.135
& 0.184
& 0.408
& 0.408 \\

\rowcolor{blue!5}
\textbf{SCOPE-RL}
& 3
& \textbf{0.403}
& \textbf{0.225}
& 0.407
& \textbf{0.368} \\

\midrule

GRPO
& 4
& 0.118
& 0.224
& 0.378
& 0.398 \\

\rowcolor{blue!5}
\textbf{SCOPE-RL}
& 4
& \textbf{0.430}
& \textbf{0.265}
& \textbf{0.427}
& \textbf{0.318} \\

\bottomrule
\end{tabular}
\end{table}

Under GRPO, entropy continuously decreases, while the proportion of all-correct prompts increases mainly at the expense of mixed prompts and the proportion of all-wrong prompts changes only slightly. In contrast, SCOPE-RL substantially increases the proportion of all-correct prompts while reducing all-wrong prompts, and maintains a higher proportion of mixed prompts at the end of training. These results suggest that the preserved entropy is associated with continued discovery of successful reasoning trajectories rather than merely increased output randomness.

\subsection{Robustness to Clipping Settings}
\label{app:clip_robustness}

\begin{wrapfigure}{r}{0.35\textwidth}
\vspace{-3mm}
\centering
\includegraphics[width=0.35\textwidth]{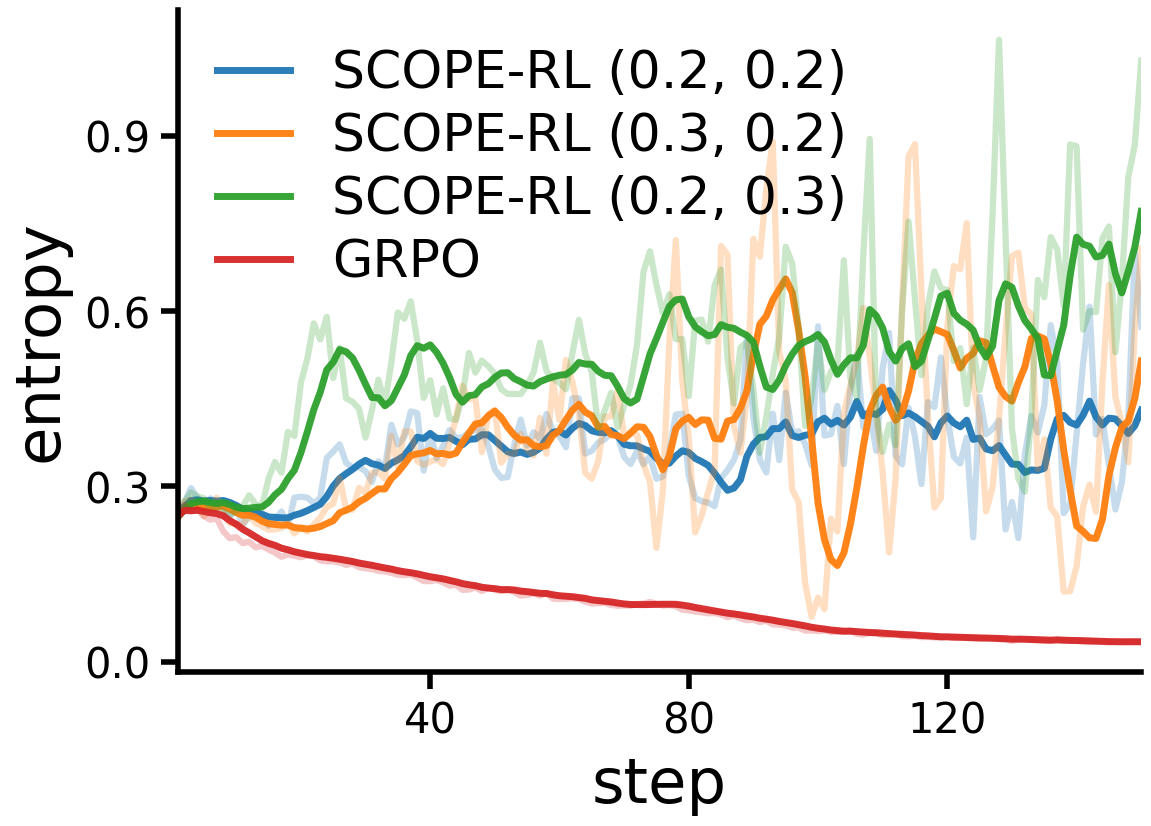}
\vspace{-2mm}
\caption{
Entropy trajectories of SCOPE-RL under different clipping settings on Qwen2.5-Math-7B. 
}
\label{fig:clip_robustness}
\vspace{-4mm}
\end{wrapfigure}
We further study whether SCOPE-RL remains effective under different clipping settings. In addition to the default symmetric clipping $(\epsilon_{\mathrm{low}}, \epsilon_{\mathrm{high}}) = (0.2, 0.2)$, we conduct controlled experiments with two asymmetric variants, $(0.2, 0.3)$ and $(0.3, 0.2)$, both combined with SCOPE-RL. Fig.~\ref{fig:clip_robustness} reports the corresponding entropy trajectories.

GRPO exhibits clear entropy collapse, with entropy decreasing from 0.2368 to 0.0395. In contrast, SCOPE-RL maintains substantially higher entropy under all three clipping configurations. In the final interval, the two asymmetric variants retain entropy values of 0.5028 and 0.5795, respectively.

These results show that changing clipping bounds affects the absolute entropy trajectory, but does not undermine the entropy control effect of SCOPE-RL. Together with the broader experiments across models, benchmarks, random seeds, and identical rollout budgets, this provides further evidence that SCOPE-RL is empirically robust to clipping settings.

\section{Detailed implementation \label{app:detail}}  
\subsection{Availability of Positive Samples in the Auxiliary Branch}
\label{app:positive_availability}

The auxiliary branch of SCOPE-RL relies on positive samples. In practice, we handle their availability as follows:

\begin{itemize}
    \item \textbf{Requirement on training difficulty.} RLVR requires the training data to provide a sufficient number of positive reward signals. In practice, the dataset should not be excessively difficult, and we generally maintain an initial pass rate above approximately $30\%$. Otherwise, insufficient positive feedback becomes a general limitation of RLVR training rather than a limitation specific to
    SCOPE-RL.

    \item \textbf{Dynamic collection of auxiliary positive samples.} In our default setting, the standard rollout batch contains $512\times 8=4096$ samples, while the auxiliary branch targets 64 positive samples. We use the pass rate from the previous training step to estimate
    the number of auxiliary prompts required to obtain approximately 64 positives and then perform the rollout. If more than 64 positive samples are obtained, we retain 64; if fewer than 64 are obtained, we use all available positive samples without additional sampling.

    \item \textbf{Robustness to insufficient positives.}
    SCOPE-RL does not require an exact number of positive samples at every training step. As shown in Fig.~\ref{fig:alpha_study}, even 32 auxiliary samples are sufficient to maintain effective entropy control. In our experiments, cases with fewer than 16 positive auxiliary samples are rare. Even if the entropy-control strength is temporarily weaker at an individual step, SCOPE-RL can recover at subsequent steps because such low-positive-rate events empirically do not persist across consecutive updates.
\end{itemize}

\subsection{Parameter settings}
We follow the default EasyR1 setup for all experiments and run all models on 8 A800 GPUs. The full training configuration is listed in Table~\ref{tab:implementation}. 

\begin{table}[t]
\centering
\caption{Implementation settings. Unless stated otherwise, the RL and SCOPE-RL settings apply to the three Qwen main experiments.}
\label{tab:implementation}
\small
\renewcommand{\arraystretch}{1.08}
\begin{tabular}{p{0.42\columnwidth}p{0.48\columnwidth}}
\toprule
\textbf{Setting} & \textbf{Value} \\
\midrule
\multicolumn{2}{l}{\textit{Qwen main experiments}} \\
Training data & DAPO-17K \\
Hardware & $8\times$ A800 GPUs (40 GB each) \\
Optimizer; learning rate & AdamW; $1\times10^{-6}$ \\
Training seed & 1 \\
Training steps & 340 (Qwen2.5-Math-7B); 204 (Qwen2.5-7B); 170 (Qwen3-4B) \\
Prompt batch size; mini-batch size & 512; 128 \\
Rollouts per prompt & 8 ($512\times8=4096$ standard rollouts per batch) \\
Rollout temperature; maximum response length & 1.0; 8192 tokens \\
Clipping bounds $(\epsilon_{\mathrm{low}},\epsilon_{\mathrm{high}})$ & $(0.2,0.2)$ \\
Reward & Binary, verifiable reward \\
\midrule
\multicolumn{2}{l}{\textit{Llama-3.1-8B-Instruct extension}} \\
Training data and duration & Math12K; four epochs (91 updates) \\
Training seeds & 1, 2, and 3 \\
Rollout-budget control & Standard rollouts reduced by the number of auxiliary rollouts; equal total rollout count across methods \\
\midrule
\multicolumn{2}{l}{\textit{SCOPE-RL auxiliary branch}} \\
Target entropy $\mathcal{H}_0$ & 0.5 \\
Temperature range & $[0.8,1.2]$ \\
Nominal coefficient $\alpha$ & $64/(512\times8)=1/64$ \\
Target positive trajectories & 64 retained per update; use all available if fewer are found \\
Auxiliary candidate count & Estimated from the preceding update's pass rate \\
Feedback statistic & Mean token entropy over valid generated tokens in the preceding update \\
\midrule
\multicolumn{2}{l}{\textit{Evaluation}} \\
Maximum response length; top-$p$ & 8192 tokens; 0.95 \\
Sampling temperature & 0.1 for Pass@1; 1.0 for Pass@$k$ \\
\bottomrule
\end{tabular}
\end{table}

\section{Proof of Theorem \ref{theo:entropy}}\label{app:key theorem}
\subsection{Preliminary}
We consider the problem of fine-tuning Large Language Models using Reinforcement Learning within a mathematically rigorous framework. The LLM generation process is modeled as a finite-horizon Markov Decision Process.
\paragraph{Notations} 
Let $Q$ denote the query space, and let $P(Q)$ be the distribution over queries. Let $\mathcal{V}$ denote the discrete token space (vocabulary). Given a query $q \sim P(Q)$, the model generates a response trajectory $o = (o_{1}, o_{2}, \dots, o_{|o|})$, where each token $o_{t} \in \mathcal{A}$ and the sequence length $|o|$ is bounded by a maximum length $L$. To ensure a unified mathematical formulation, we assume all trajectories have a fixed length $L$. For any response shorter than $L$, we append a special padding token $o_{\text{pad}}$ to its suffix until the length reaches $L$.
At time step $t$, the state $s_{t}$ is defined as the concatenation of the query and the partial response generated so far, i.e., $s_{t} = (q, o_{<t})$. The policy $\pi_{\theta}(\cdot | s_{t})$, parameterized by $\theta$, maps the current state $s_{t}$ to a probability distribution over $\mathcal{A}$. The probability of generating a complete trajectory $o$ given $q$ is given by the chain rule:
\[\small  \pi_{\theta}(o|q) = \prod_{t=1}^{L} \pi_{\theta}(o_{t} | s_{t}).\]

Throughout this section, we adopt the same definitions and notation used in the main text. We formally define entropy at two different levels of granularity.

First, we define the token-level entropy at time step $t$ given state $s_{t}$ as the Shannon entropy of the policy's output distribution:
\begin{equation}\label{eq:token level entropy}
    \small
    \begin{aligned}
        \mathcal{H}_{t}(\pi_{\theta}; s_{t}) \coloneqq -\sum_{o_{t} \in \mathcal{A}} \pi_\theta(o_{t} | s_t) \log \pi_\theta(o_{t} | s_{t}) = -\mathbb{E}_{o_{t} \sim \pi_\theta(\cdot|s_t)} \left[\log \pi_\theta(a | s_{t})\right].
    \end{aligned}   
\end{equation}
Based on this, we define the cumulative expected entropy $\mathcal{H}(\pi_{\theta}; q)$ for a query $q$ as the expectation of the sum of token-level entropy $\mathcal{H}_{t}$ over the generated trajectory:
\begin{equation}\label{eq:cumulative token level entropy}
    \small
    \begin{aligned}
        \mathcal{H}(\pi_{\theta}; q) \coloneqq \mathbb{E}_{o \sim \pi_{\theta}(O|q)} \left[ \sum_{t=1}^{L} \mathcal{H}_{t}(\pi_\theta; s_{t}) \right].
    \end{aligned}
\end{equation}

Second, we define the trajectory-level entropy $\mathcal{H}^{\prime}(\pi_{\theta}; q)$, which measures the uncertainty of the entire response sequence $o$:
\begin{equation}\label{eq:trajectory level entropy}
    \small
    \begin{aligned}
        \mathcal{H}^{\prime}(\pi_{\theta}; q) \coloneqq -\sum_{o}\pi_{\theta}(o|q) \cdot \log \pi_{\theta}(o|q)= -\mathbb{E}_{o \sim \pi_{\theta}(O|q)} \left[ \log \pi_{\theta}(o | q) \right].
    \end{aligned}
\end{equation}

The following lemma establishes the equivalence between these two formulations $\mathcal{H}(\pi_{\theta}; q)$ and $\mathcal{H}^{\prime}(\pi_{\theta}; q)$, providing the theoretical basis for optimizing sequence diversity via step-wise regularization.
\begin{lemma}\label{lem:empirical entropy is trajectory level entropy}
    The cumulative expected token-level entropy $\mathcal{H}(\pi_{\theta}; q)$ is equivalent to the trajectory-level entropy $\mathcal{H}^{\prime}(\pi_{\theta}; q)$. That is, for any policy $\pi_{\theta}$ and query $q$, it holds that
    \begin{equation*}
        \small
        \mathcal{H}(\pi_{\theta};q)=\sum_{t=1}^{L}\mathbb{E}_{s_{t}\sim\pi_{\theta}(\cdot|q)}[\mathcal{H}_{t}(\pi_{\theta};s_{t})]=\mathcal{H}^{\prime}(\pi_{\theta}; q).
    \end{equation*}
\end{lemma}
\begin{proof}
    We begin by considering the cumulative expected token-level entropy. According to \eqref{eq:token level entropy} and \eqref{eq:cumulative token level entropy}, it holds that
    \begin{equation}\label{eq:token level entropy analysis}
        \small
        \begin{aligned}
            \mathcal{H}(\pi_{\theta};q)= & \mathbb{E}_{o \sim \pi_{\theta}(O|q)}\left[\sum_{t=1}^{L}\mathcal{H}_{t}(\pi_{\theta};s_{t})\right]
            =\sum_{t=1}^{L}\mathbb{E}_{o\sim \pi_{\theta}(O|q)}[\mathcal{H}_{t}(\pi_{\theta};s_{t})]
            =\sum_{t=1}^{L}\mathbb{E}_{s_{t}\sim\pi_{\theta}(\cdot|q)}[\mathcal{H}_{t}(\pi_{\theta};s_{t})]
        \end{aligned}
    \end{equation}
    Next, we turn to the analysis of the trajectory-level entropy, as defined in \eqref{eq:trajectory level entropy}.
    \begin{equation}\label{eq:trajectory level entropy analysis}
        \small
        \begin{aligned}
            \mathcal{H}^{\prime}(\pi_{\theta}; q) = & -\mathbb{E}_{o \sim \pi_{\theta}(O|q)} \left[ \log \pi_{\theta}(o | q) \right]
            =-\mathbb{E}_{o \sim \pi_{\theta}(O|q)} \left[ \sum_{t=1}^{L}\log \pi_{\theta}(o_{t} | s_{t}) \right]
            \\ = & -\sum_{t=1}^{L}\mathbb{E}_{o \sim \pi_{\theta}(O|q)} \left[ \log \pi_{\theta}(o_{t} | s_{t}) \right]
            =-\sum_{t=1}^{L}\mathbb{E}_{s_{t+1} \sim \pi_{\theta}(\cdot|q)} \left[ \log \pi_{\theta}(o_{t} | s_{t}) \right]
            \\ = & \sum_{t=1}^{L}\mathbb{E}_{s_{t} \sim \pi_{\theta}(\cdot|q)}[\mathcal{H}_{t}(\pi_{\theta};s_{t})].
        \end{aligned}
    \end{equation}
    Combining \eqref{eq:token level entropy analysis} and \eqref{eq:trajectory level entropy analysis}, we prove the conclusion.
\end{proof}
\begin{remark}
    Lemma \ref{lem:empirical entropy is trajectory level entropy} justifies the practical implementation of entropy regularization in auto-regressive models. The llama establishes an equality between trajectory entropy and the expected sum of conditional token entropies under the fixed-length formulation. In practice, we monitor the mean entropy over valid generated tokens to reduce sensitivity to response length. This length-normalized statistic is an empirical control signal; when response lengths vary, it is not generally equal to trajectory entropy.

\end{remark}

\subsection{Problem Simplified}\label{app:sentence to token}
Exploration in LLM reasoning is intuitively reflected by the diversity of the generated response trajectories. Following Lemma \ref{lem:empirical entropy is trajectory level entropy}, this global diversity is mathematically equivalent to the cumulative token-level entropy $\mathcal{H}(\pi_{\theta};q)$. We therefore use mean token entropy as a practical proxy for policy uncertainty during training, while the equivalence in Lemma 1 applies to expected cumulative token entropy under its stated assumptions.

In this work, we analyze the relationship between entropy dynamics and parameter updates. While the reinforcement learning objective targets sequence-level attributes, the training process physically operates on the shared parameters $\theta$ at each specific token generation step. Based on the linearity of differentiation, the gradient of the sequence entropy is fundamentally driven by the superposition of gradients from individual time steps:
\begin{equation}
    \small
    \label{eq:entropy decomposition}
    \begin{aligned}
        \nabla_{\theta} \mathcal{H}(\pi_{\theta}; q) & =\nabla_{\theta}\mathbb{E}_{o \sim \pi_{\theta}(O|q)} \left[ \sum_{t=1}^{L} \mathcal{H}_{t}(\pi_\theta; s_{t}) \right]
        \\ & =\mathbb{E}_{o \sim \pi_{\theta}(O|q)}\left[\sum_{t=1}^{L}\left(\nabla_{\theta}\mathcal{H}_{t}(\pi_{\theta};q)+\mathcal{H}_{t}(\pi_{\theta};q)\cdot\nabla_{\theta}\log \pi_{\theta}(o|q)\right)\right] 
        \\ & \approx \mathbb{E}_{o \sim \pi_{\theta}(O|q)} \left[ \sum_{t=1}^{L} \nabla_\theta \mathcal{H}_{t}(\pi_\theta; s_t) \right].
    \end{aligned}
\end{equation}


\begin{remark}
As explicitly derived in \eqref{eq:entropy decomposition}, the full gradient of the sequence entropy comprises two distinct components: (1) the instantaneous gradient term $\nabla_{\theta} \mathcal{H}_{t}$, which reflects how the parameter update directly alters the shape of the policy distribution at the current state; and (2) the trajectory-shift term (the positive-sample regularizer weighted by cumulative entropy), which accounts for how policy changes alter the distribution of states likely to be visited in future steps. In the subsequent theoretical analysis, we explicitly focus on the instantaneous gradient component. This theoretical simplification is justified because our primary goal is to uncover the micro-mechanism of how temperature modulation re-allocates the gradient contribution among tokens (particularly by dampening the excessive reinforcement of dominant tokens) at any given decision step, thereby locally and directly counteracting the tendency of the policy to collapse toward determinism. While trajectory shift is relevant for long-term value propagation, the modification of the local distributional shape constitutes the dominant effect regarding the prevention of entropy collapse. Therefore, isolating this component allows us to mathematically decouple and rigorously prove the direct control authority of temperature $T$ over the optimization dynamics.
\end{remark}

\subsection{Detailed Proof}
In this section, we provide the rigorous proof for the temperature-modulated entropy control mechanism. As justified in Section \ref{app:sentence to token}, we focus on a contextual bandit setting.
\subsubsection{Problem Setup and Definition}
Consider a generic state $s$, we denote the action space as $\mathcal{A}$. The policy $\pi_{\theta}(\cdot|s)$ is parameterized as a Softmax distribution over the action space $\mathcal{A}$:
\begin{align*}
\small
    \pi_{\theta}(a|s)=\frac{\exp(z(s,a))}{\sum_{a' \in \mathcal{A}}\exp(z(s,a'))},
\end{align*}
where $z(s,a)$ is the logit for the state-action pair $(s,a)$ under parameter $\theta$. Furthermore, given a temperature T, we define:
\begin{align*}
\small
    \pi_{\theta}^{T}(a|s)=\frac{\exp(\frac{z(s,a)}{T})}{\sum_{a' \in \mathcal{A}}\exp(\frac{z(s,a')}{T})},
\end{align*}
we write $z_{\theta_{k}}(s,a)$ simply as $z_{k}(s,a)$, and we write $\pi_{\theta}^{T}(a|s)$ simply as $\pi_{\theta}(a|s)$ when $T=1$.

The standard expected-reward objective for this contextual bandit problem is
\begin{equation}
	\label{eq:standard-bandit-objective}
	\small
	J(\theta)
	=
	\mathbb{E}_{a\sim\pi_{\theta}(\cdot\mid s)}
	\left[R(s,a)\right].
\end{equation}
where $R(s,a)$ is the reward for token $a$ given positive context $s$.

To isolate the effect of temperature-modulated sampling, at iteration $k$ we
consider the following auxiliary surrogate objective with the sampling distribution
$\pi_{\theta_k}^{T}$ held fixed:
\begin{equation}
	\label{eq:aux-bandit-surrogate}
	\small
	J_{\mathrm{aux}}^{(k)}(\theta;T)
	=
	\mathbb{E}_{a\sim\pi_{\theta_k}^{T}(\cdot\mid s)}
	\left[
	R(s,a)\log\pi_{\theta}(a\mid s)
	\right].
\end{equation}
The corresponding local update is
\begin{equation}
	\small
	\theta_{k+1}
	=
	\theta_k
	+
	\left.
	\eta\nabla_{\theta}
	J_{\mathrm{aux}}^{(k)}(\theta;T)
	\right|_{\theta=\theta_k}
	=
	\theta_k
	+
	\eta
	\mathbb{E}_{\textcolor{red}{a \sim \pi_{\theta_{k}}^{T}(\cdot | s)}}
	\left[
	R(s,a)\nabla_{\theta}
	\log\pi_{\theta_k}(a\mid s)
	\right].
	\label{eq:positive-sample update simplify}
\end{equation}
where $\eta$ is the learning rate.
\begin{remark}
	Note that, unlike the standard RL policy gradient, \eqref{eq:positive-sample update simplify} estimates the policy gradient using samples drawn from the fixed temperature-modulated policy $\pi_{\theta_k}^{T}$. This mismatch introduces a deliberate bias into the gradient estimate. SCOPE-RL is designed to exploit this temperature-induced bias, and the following analysis characterizes how it reshapes the policy update to regulate policy entropy.
\end{remark}
Within this contextual bandit framework, we define the policy entropy for parameter $\theta$ at a generic state $s$ a:
\begin{equation}
    \small
    \mathcal{H}(\pi_{\theta})=-\mathbb{E}_{a \sim \pi_{\theta}(\cdot|s)}[\log \pi_{\theta}(a|s)]
\end{equation}
\subsubsection{Theoretical Assumptions}
To render the analysis tractable and reveal the core mechanism, we introduce the following standard assumptions.

\begin{assumption}[Binary Reward]
    For a given initial state $s$, a corresponding action space $\mathcal{A}$, the reward for $(s,a)$ is binary:
    \begin{align*}
        \small
        R(s,a)=
        \begin{cases}
        1\quad \text{if }\ a \in \mathcal{A}^{*};\\
        0\quad \text{otherwise},
        \end{cases}
    \end{align*}
    where $\mathcal{A}^{*}$ denotes the set of reference actions for state $s$.
\end{assumption}

\begin{assumption}[Orthogonal Gradients]\label{ass:orthogonal_gradients}
    Following standard analyses in deep learning theory (e.g., NTK regime), we assume the gradients of logits for different tokens are orthogonal, i.e., 
    \begin{equation*}
        \small
        \left\langle\nabla_{\theta}z_{k}(s,a),\nabla_{\theta}z_{k}(s,b)\right\rangle=c_{a,b} \cdot \delta_{a,b},\quad \forall a,b\in \mathcal{A},
    \end{equation*}  
    where $c_{a,b} \in \mathbb{R}$ and $\delta_{a,b}=\textbf{1}[a=b]$ is the Kronecker delta. In the following, we set $c_{a,b} \equiv 1$.
\end{assumption}

\begin{remark}
We acknowledge that Assumption \ref{ass:orthogonal_gradients} is an idealization given the shared parameters in LLMs. However, assuming gradient orthogonality is a standard practice in deep learning theory (e.g., the Neural Tangent Kernel regime \cite{jacot2018neural}) to model the capacity of over-parameterized networks to adjust outputs independently. Crucially, this assumption decouples the complex interactions between tokens, allowing us to mathematically isolate the mechanistic effect of temperature modulation on the gradient weights, thereby rendering the qualitative analysis tractable.
\end{remark}

\subsubsection{Formal statement of Theorem \ref{theo:entropy}}
\begin{theorem}
	\label{prof:temperature}
	Let the sequence of parameters $\{\theta_k\}$ be generated by
	\eqref{eq:positive-sample update simplify}. Define
	\begin{align}
		\Delta\mathcal{H}_k(T)
		=
		-\eta
		\sum_{a^*\in\mathcal{A}^*}
		\pi_{\theta_k}^{T}(a^*\mid s)
		\operatorname{Cov}_{\pi_{\theta_k}}
		\left(
		z_k(s,a'),
		\delta_{a',a^*}-\pi_{\theta_k}(a'\mid s)
		\right).
		\label{eq:deltaH-definition}
	\end{align}
	Then, under the first-order approximation, 
	\begin{align}
		\Delta\mathcal{H}_k(T)
		\approx
		\mathcal{H}(\pi_{\theta_{k+1}})
		-
		\mathcal{H}(\pi_{\theta_k}).
		\label{eq:deltaH-approx}
	\end{align}
	Moreover, if
	\begin{align*}
		\Delta\mathcal{H}_k(1)<0,
	\end{align*}
	then
	\begin{align}
		\left.
		\frac{\mathrm{d}\Delta\mathcal{H}_k(T)}
		{\mathrm{d}T}
		\right|_{T=1}
		>0.
		\label{eq:strict-temperature-derivative}
	\end{align}
\end{theorem}
\begin{proof}
    For a given parameter $\theta_{k}$ and a relatively small learning rate $\eta$, leveraging Taylor'expansion under first-order approximation, we have 
    \begin{align*}
        \small
        \mathcal{H}(\pi_{\theta_{k+1}})- \mathcal{H}(\pi_{\theta_{k}}) \approx \langle\nabla_{\theta} \mathcal{H}(\pi_{\theta_{k}}), \theta_{k+1}-\theta_{k}\rangle.
    \end{align*}
    We then turn to derive what $\nabla_{\theta} \mathcal{H}(\pi_{\theta_{k}})$ is, according to the definition of $\mathcal{H}$, we have
    \begin{equation*}
        \small
        \begin{aligned}
        \nabla_{\theta} \mathcal{H}(\pi_{\theta_{k}})& =-\nabla_{\theta}\mathbb{E}_{a\sim \pi_{\theta_{k}}(\cdot | s)}[\log\pi_{\theta_{k}}(a|s)]
        \\ &=-\nabla_{\theta}\sum_{a \in \mathcal{A}}[\pi_{\theta_{k}}(a|s) \cdot \log\pi_{\theta_{k}}(a|s)]
        \\ &=-\mathbb{E}_{a \sim \pi_{\theta_{k}}}[\nabla_{\theta}\log\pi_{\theta_{k}}(a|s) \cdot \log\pi_{\theta_{k}}(a|s)+\nabla_{\theta}\log\pi_{\theta_{k}}(a|s)]
        \\ & =-\mathbb{E}_{a \sim \pi_{\theta_{k}}}[\nabla_{\theta}\log\pi_{\theta_{k}}(a|s) \cdot \log\pi_{\theta_{k}}(a|s)].
    \end{aligned}
    \end{equation*}
    the last equality follows from $\small \sum_{a \in \mathcal{A}}\nabla_{\theta}\pi_{\theta_{k}}(a|s)=\nabla_{\theta}\sum_{a \in \mathcal{A}}\pi_{\theta_{k}}(a|s)=\nabla_{\theta}1=0$.
    
    Then we have
    \begin{align*}
    \small
        \mathcal{H}(\pi_{\theta_{k+1}})- \mathcal{H}(\pi_{\theta_{k}}) &\approx -\left\langle\mathbb{E}_{a \sim \pi_{\theta_{k}}(\cdot |s)}[\nabla_{\theta}\log\pi_{\theta_{k}}(a|s) \cdot \log\pi_{\theta_{k}}(a|s)], \theta_{k+1}-\theta_{k}\right\rangle
        \\ &=-\mathbb{E}_{a \sim \pi_{\theta_{k}}(\cdot | s)}\left[\log\pi_{\theta_{k}}(a|s)\cdot \langle\nabla_{\theta}\log\pi_{\theta_{k}}(a|s),\theta_{k+1}-\theta_{k}\rangle\right]
        \\ & \approx -\mathbb{E}_{a \sim \pi_{\theta_{k}}(\cdot | s)}\left[\log\pi_{\theta_{k}}(a|s)\cdot (\log\pi_{\theta_{k+1}}(a|s)-\log\pi_{\theta_{k}}(a|s))\right] 
        \\ & \approx -\mathbb{E}_{a \sim \pi_{\theta_{k}}(\cdot | s)}\left[\log\pi_{\theta_{k}}(a|s)\cdot \sum_{a' \in \mathcal{A}}\frac{\textbf{d} \log\pi_{\theta_{k}}(a|s)}{\textbf{d}z_{k}(s,a')}\cdot \Delta z_{k}(s,a')\right].
    \end{align*}
    We now evaluate $\frac{\textbf{d} \log\pi_{\theta_{k}}(a|s)}{\textbf{d}z_{k}(s,a')}$ and $\Delta z_{k}(s,a')$ in turn.
    \begin{align*}
    \small
        \frac{\textbf{d} \log\pi_{\theta_{k}}(a|s)}{\textbf{d}z_{k}(s,a')}= \delta_{a,a'}-\pi_{\theta_{k}}(a'|s) ;
    \end{align*}
    According to the parameter update rule in \eqref{eq:positive-sample update simplify}, it holds that
    \begin{equation}\label{eq:delta z}
    \small
        \begin{aligned}
        \Delta z_{k}(s,a') &\approx \left\langle\frac{\textbf{d}z_{k}(s,a')}{\textbf{d}\theta},\theta_{k+1}-\theta_{k}\right\rangle 
        \\ &=\left\langle\nabla_{\theta}z_{k}(s,a'),\eta\cdot \mathbb{E}_{\textcolor{red}{a \sim \pi_{\theta_{k}}^{T}(\cdot | s)}}[\mathbf{1}[a\in\mathcal{A}^{*}] \cdot \nabla_{\theta}\log\pi_{\theta_{k}}(a|s)]\right\rangle
        \\ &=\eta \cdot \sum_{a^{*} \in \mathcal{A}^{*}}\pi_{\theta_{k}}^{T}(a^{*}|s)\cdot\left\langle\nabla_{\theta}z_{k}(s,a'),\nabla_{\theta}z_{k}(s,a^{*})-\mathbb{E}_{a \sim \pi_{\theta_{k}}}[\nabla_{\theta}z_{k}(s,a)]\right\rangle 
        \\ &=\eta \cdot \sum_{a^{*} \in \mathcal{A}^{*}}\left[\pi_{\theta_{k}}^{T}(a^{*}|s)\cdot (\delta_{a',a^{*}}-\pi_{\theta_{k}}(a'|s))\right],
    \end{aligned}
    \end{equation}
    where the first equality in \eqref{eq:delta z} follows from
    \begin{align*}
        \small
        \nabla_{\theta}\log\pi_{\theta_{k}}(a|s) &=\nabla_{\theta}\left[z_{k}(s,a)-\log\sum_{a'\in \mathcal{A}} \exp(z_{k}(s,a')) \right]
        \\ &=\nabla_{\theta}z_{k}(s,a)-\frac{1}{\sum_{a'\in \mathcal{A}} \exp(z_{k}(s,a'))}\left(\sum_{a'\in \mathcal{A}}\exp(z_{k}(s,a')) \cdot \nabla_{\theta}z_{k}(s,a')\right)
        \\ &=\nabla_{\theta}z_{k}(s,a)-\mathbb{E}_{a \sim \pi_{\theta_{k}}}[\nabla_{\theta}z_{k}(s,a)],
    \end{align*}
    and the last equality in \eqref{eq:delta z} follows from Assumption \ref{ass:orthogonal_gradients}.
    Then, it holds that
    \begin{equation}\label{eq:delta entropy 1}
    \small
        \begin{aligned}
        \mathcal{H}(\pi_{\theta_{k+1}})- \mathcal{H}(\pi_{\theta_{k}}) &\approx -\sum_{a' \in \mathcal{A}} \Delta z_{k}(s,a') \cdot \mathbb{E}_{a \sim \pi_{\theta_{k}}(\cdot | s)}\left[\log\pi_{\theta_{k}}(a|s) \cdot \frac{\textbf{d} \log\pi_{\theta_{k}}(a|s)}{\textbf{d}z_{k}(s,a')}\right] 
        \\ &=-\sum_{a' \in \mathcal{A}} \Delta z_{k}(s,a') \cdot \mathbb{E}_{a \sim \pi_{\theta_{k}}(\cdot | s)}\left[\log\pi_{\theta_{k}}(a|s) \cdot (\delta_{a,a'}-\pi_{\theta_{k}}(a'|s))\right] 
        \\ &=-\sum_{a' \in \mathcal{A}} \Delta z_{k}(s,a') \cdot\left[\pi_{\theta_{k}}(a'|s)\cdot \log\pi_{\theta_{k}}(a'|s)-\pi_{\theta_{k}}(a'|s)\cdot \mathbb{E}_{a\sim \pi_{\theta_{k}}(\cdot | s)}[\log\pi_{\theta_{k}}(a|s)]\right]
        \\ &=-\left(\mathbb{E}_{a\sim \pi_{\theta_{k}}(\cdot | s)}[\Delta z_{k}(s,a) \cdot \log\pi_{\theta_{k}}(a|s)]-\mathbb{E}_{a\sim \pi_{\theta_{k}}(\cdot | s)}[\Delta z_{k}(s,a)]\cdot \mathbb{E}_{a\sim \pi_{\theta_{k}}(\cdot | s)}\left[\log\pi_{\theta_{k}}(a|s)\right]\right)
        \\ &=-\text{Cov}_{\pi_{\theta_{k}}}\left(\Delta z_{k}(s,a),\log\pi_{\theta_{k}}(a|s)\right) 
        \\ &=-\text{Cov}_{\pi_{\theta_{k}}}(\Delta z_{k}(s,a),z_{k}(a|s)) ,
    \end{aligned}
    \end{equation}
    take \eqref{eq:delta z} into \eqref{eq:delta entropy 1}, we have
    \begin{align*}
    \small
        \mathcal{H}(\pi_{\theta_{k+1}})- \mathcal{H}(\pi_{\theta_{k}}) &\approx -\eta \cdot \sum_{a^{*} \in \mathcal{A}^{*}} \pi_{\theta_{k}}^{T}(a^{*}|s)\cdot \text{Cov}_{\pi_{\theta_{k}}}(z_{k}(s,a),\delta_{a,a^{*}}-\pi_{\theta_{k}}(a|s)) \equiv \Delta \mathcal{H}_{k}(T).              
    \end{align*} 
    We now proceed to the proof of the second part of the lemma. We first partition the set $\mathcal{A}^{*}$ into three subsets: $\mathcal{A}_{1}$, $\mathcal{A}_{2}$, and $\mathcal{A}_{3}$, where
    \begin{align}\label{eq:three subset}
    \small
        &\mathcal{A}_{1}=\left\{a|a\in \mathcal{A}^{*} , z_{k}(s,a)<\mathbb{E}_{a'\sim \pi_{\theta_{k}}(\cdot | s)}[z(s,a')]\right\};
        \\ &\mathcal{A}_{2}=\left\{a|a\in \mathcal{A}^{*} , z_{k}(s,a) \ge \mathbb{E}_{a'\sim \pi_{\theta_{k}}(\cdot | s)}[z(s,a')], \text{Cov}_{a' \sim \pi_{\theta_{k}}}(z_{k}(s,a'),\delta_{a',a}) < C_{\pi_{k}}\right\};
        \\ &\mathcal{A}_{3}=\left\{a|a\in \mathcal{A}^{*} ,  z_{k}(s,a) \ge \mathbb{E}_{a'\sim \pi_{\theta_{k}}(\cdot | s)}[z(s,a')], \text{Cov}_{a' \sim \pi_{\theta_{k}}}(z_{k}(s,a'),\delta_{a',a}) \ge C_{\pi_{k}}\right\};
    \end{align}
    where $C_{\pi_{k}}=\text{Cov}_{a' \sim \pi_{\theta_{k}}}(z_{k}(s,a'),\pi_{\theta_{k}}(a'|s))$.
    According to the condition $\Delta\mathcal H_k(1)<0$, we have
    \begin{align}
    	\label{eq:strict-positive-sum}
    	\small
        0 &< \sum_{a^{*}\in \mathcal{A}^{*}}\pi_{\theta_{k}}(a^{*}|s)\cdot \text{Cov}_{\pi_{\theta_{k}}}(z_{k}(s,a),\delta_{a,a^{*}}-\pi_{\theta_{k}}(a|s)) \nonumber 
        \\ &=\sum_{i=1}^{3}\sum_{a^{*}\in \mathcal{A}_{i}}\pi_{\theta_{k}}(a^{*}|s) \cdot \text{Cov}_{\pi_{\theta_{k}}}(z_{k}(s,a),\delta_{a,a^{*}}-\pi_{\theta_{k}}(a|s)) \nonumber
        \\ &\le \sum_{a^{*}\in \mathcal{A}_{2}\bigcup \mathcal{A}_{3}}\pi_{\theta_{k}}(a^{*}|s) \cdot \text{Cov}_{\pi_{\theta_{k}}}(z_{k}(s,a),\delta_{a,a^{*}}-\pi_{\theta_{k}}(a|s)).
    \end{align}
    We now turn to the analysis of $ \frac{\textbf{d}\Delta \mathcal{H}_{k}(T)}{\textbf{d}T}$. According to the definition of $\pi_{\theta}$, it holds that
    \begin{align*}
    \small
        \frac{\textbf{d}\pi_{\theta_{k}}^{T}(a|s)}{\textbf{d}T} 
         = -T^{-2}\cdot \pi_{\theta_{k}}^{T}(a|s)\cdot \left[z_{k}(s,a)-\mathbb{E}_{\pi_{\theta_{k}}^{T}}[z_{k}(s,a')]\right].
    \end{align*}
    We evaluate the temperature derivative at the reference temperature $T=1$.
    At $T=1$, we have
    \begin{align*}
    \small
        \frac{\textbf{d}\pi_{\theta_{k}}^{T}(a|s)}{\textbf{d}T}|_{T=1} =-\pi_{\theta_{k}}(a|s)\cdot \left[z_{k}(s,a)-\mathbb{E}_{\pi_{\theta_{k}}}[z_{k}(s,a')]\right] =-\text{Cov}_{\pi_{\theta_{k}}}(z_{k}(s,a'),\delta_{a,a'}).
    \end{align*}
    Thus it holds that
    \begin{align*}
    \small
        \frac{\textbf{d}\Delta \mathcal{H}(T)}{\textbf{d}T}|_{T=1} &=-\eta \cdot \sum_{a^{*}\in \mathcal{A}^{*}}\text{Cov}_{\pi_{\theta_{k}}}(z_{k}(s,a'),\delta_{a',a^{*}}-\pi_{\theta_{k}}(a'|s)) \cdot \frac{\textbf{d}\pi_{\theta_{k}}^{T}(a^{*}|s)}{\textbf{d}T}
        \\ &=-\eta \cdot \sum_{a^{*}\in \mathcal{A}^{*}}\text{Cov}_{\pi_{\theta_{k}}}(z_{k}(s,a'),\delta_{a',a^{*}}-\pi_{\theta_{k}}(a'|s)) \cdot (-\text{Cov}_{\pi_{\theta_{k}}}(z_{k}(s,a'),\delta_{a',a^{*}}))
        \\ &=\eta \cdot \left[\sum_{a^{*}\in \mathcal{A}_{1}\bigcup \mathcal{A}_{2} \bigcup \mathcal{A}_{3}}\text{Cov}_{\pi_{\theta_{k}}}(z_{k}(s,a'),\delta_{a',a^{*}}-\pi_{\theta_{k}}(a'|s)) \cdot \text{Cov}_{\pi_{\theta_{k}}}(z_{k}(s,a'),\delta_{a',a^{*}})\right].
    \end{align*}
    We now proceed with a case-by-case analysis based on different $a^{*}$. 
    
    For $a^{*} \in \mathcal{A}_{1}$, we have $z_{k}(s,a)<\mathbb{E}_{a'\sim \pi_{\theta_{k}}(\cdot | s)}[z(s,a')]$, and then $\text{Cov}_{\pi_{\theta_{k}}}(z_{k}(s,a'),\delta_{a',a^{*}})<0$. According to Lemma \ref{lem:C-positive}, it holds that $C_{\pi_k}\geq 0$, then we have
    \begin{align}
    	\label{eq:A1 contribution}
    	\text{Cov}_{\pi_{\theta_{k}}}(z_{k}(s,a'),\delta_{a',a^{*}}-\pi_{\theta_{k}}(a'|s))=\text{Cov}_{\pi_{\theta_{k}}}(z_{k}(s,a'),\delta_{a',a^{*}})-C_{\pi_k}<0,
    \end{align} 
    therefore
    \begin{align}
    	\label{eq:A1 contribution derivative}
    	\small
        \sum_{a^{*} \in \mathcal{A}_{1}}\text{Cov}_{\pi_{\theta_{k}}}(z_{k}(s,a'),\delta_{a',a^{*}}-\pi_{\theta_{k}}(a'|s)) \cdot \text{Cov}_{\pi_{\theta_{k}}}(z_{k}(s,a'),\delta_{a',a^{*}})\ge 0,
    \end{align}
    We first consider the boundary case that $\mathcal{A}_{2}=\varnothing$, by combining \eqref{eq:strict-positive-sum} and \eqref{eq:A1 contribution}, we have
    \begin{align}
    	\sum_{a^*\in\mathcal A_3}
    	\pi_{\theta_k}(a^*\mid s)
    	\operatorname{Cov}_{\pi_{\theta_k}}
    	\left(
    	z_k(s,a),
    	\delta_{a,a^*}-\pi_{\theta_k}(a\mid s)
    	\right)
    	>0.
    	\label{eq:A3-positive-when-A2-empty}
    \end{align}
    Therefore, there exists at least one $a^*\in\mathcal A_3$ such that
    \begin{align}
    	\operatorname{Cov}_{\pi_{\theta_k}}
    	\left(
    	z_k(s,a),
    	\delta_{a,a^*}-\pi_{\theta_k}(a\mid s)
    	\right)
    	>0.
    \end{align} 
    For the particular $a^*\in\mathcal A_3$ whose existence is guaranteed
    by \eqref{eq:A3-positive-when-A2-empty}, we further have
    \begin{align}
    	\operatorname{Cov}_{\pi_{\theta_k}}\left(z_k(s,a'),\delta_{a',a^*}\right)
    	=
    	\operatorname{Cov}_{\pi_{\theta_k}}
    	\left(z_k(s,a'),\delta_{a',a^*}-\pi_{\theta_k}(a'\mid s)\right)
    	+C_{\pi_k}
    	>C_{\pi_k}
    	\geq 0.
    	\label{eq:A3-second-cov-positive}
    \end{align}
    Therefore, the product corresponding to this $a^*$ is strictly positive,
    while all other terms in $\mathcal A_3$ are non-negative. Hence,
    \begin{align}
    	\sum_{a^*\in\mathcal A_3}
    	\operatorname{Cov}_{\pi_{\theta_k}}
    	\left(
    	z_k(s,a'),
    	\delta_{a',a^*}
    	-
    	\pi_{\theta_k}(a'\mid s)
    	\right)\cdot
    	\operatorname{Cov}_{\pi_{\theta_k}}
    	\left(
    	z_k(s,a'),
    	\delta_{a',a^*}
    	\right)
    	>0.
    	\label{eq:A3-strict-positive-contribution}
    \end{align}
    Hence the contribution from $\mathcal A_3$ to
    $\left.\frac{d\Delta\mathcal H_k(T)}{dT}\right|_{T=1}$ is strictly
    positive, while the contribution from $\mathcal A_1$ is non-negative, which is calculate in \eqref{eq:A1 contribution derivative}.
    Therefore,
    \begin{align}
    	\left.
    	\frac{d\Delta\mathcal H_k(T)}{dT}
    	\right|_{T=1}
    	>0.
    \end{align}
    
    In the case that $\mathcal A_2\neq\varnothing$. For $a^{*} \in \mathcal{A}_{2}$, we have $z_{k}(s,a^{*})\ge \mathbb{E}_{a'\sim \pi_{\theta_{k}}(\cdot | s)}[z(s,a')]$ and $\text{Cov}_{a' \sim \pi_{\theta_{k}}}(z_{k}(s,a'),\delta_{a',a^{*}}) < C_{\pi_{k}}$. Let $u=\max_{a^{*}\in \mathcal{A}_{2}}\{z_{k}(s,a^{*})\}$, therefore it holds that
    \begin{align*}
    \small
        &\sum_{a^{*} \in \mathcal{A}_{2}}\text{Cov}_{\pi_{\theta_{k}}}(z_{k}(s,a'),\delta_{a',a^{*}}-\pi_{\theta_{k}}(a'|s)) \cdot \text{Cov}_{\pi_{\theta_{k}}}(z_{k}(s,a'),\delta_{a',a^{*}}) 
        \\ =& \sum_{a^{*} \in \mathcal{A}_{2}}\text{Cov}_{\pi_{\theta_{k}}}(z_{k}(s,a'),\delta_{a',a^{*}}-\pi_{\theta_{k}}(a'|s)) \cdot \pi_{\theta_{k}}(a^{*}|s)\cdot \left[z_{k}(s,a^{*})-\mathbb{E}_{\pi_{\theta_{k}}}[z_{k}(s,a')]\right]
        \\ =&\sum_{a^{*} \in \mathcal{A}_{2}}\underbrace{\text{Cov}_{\pi_{\theta_{k}}}(z_{k}(s,a'),\delta_{a',a^{*}}-\pi_{\theta_{k}}(a'|s))}_{<0} \cdot \pi_{\theta_{k}}(a^{*}|s)\cdot \left[\underbrace{z_{k}(s,a^{*})-u}_{\le 0}+u-\mathbb{E}_{\pi_{\theta_{k}}}[z_{k}(s,a')]\right]
        \\ \ge &\sum_{a^{*} \in \mathcal{A}_{2}}\text{Cov}_{\pi_{\theta_{k}}}(z_{k}(s,a'),\delta_{a',a^{*}}-\pi_{\theta_{k}}(a'|s)) \cdot \pi_{\theta_{k}}(a^{*}|s)\cdot \left[u-\mathbb{E}_{\pi_{\theta_{k}}}[z_{k}(s,a')]\right].
    \end{align*} 
    For $a^{*} \in \mathcal{A}_{3}$, we have
    \begin{equation}
    \begin{split}
            \small
    \label{eq:limit}
        & \sum_{a^{*} \in \mathcal{A}_{3}}\text{Cov}_{\pi_{\theta_{k}}}(z_{k}(s,a'),\delta_{a',a^{*}}-\pi_{\theta_{k}}(a'|s)) \cdot \text{Cov}_{\pi_{\theta_{k}}}(z_{k}(s,a'),\delta_{a',a^{*}}) 
        \\ =& \sum_{a^{*} \in \mathcal{A}_{3}}\text{Cov}_{\pi_{\theta_{k}}}(z_{k}(s,a'),\delta_{a',a^{*}}-\pi_{\theta_{k}}(a'|s)) \cdot \pi_{\theta_{k}}(a^{*}|s)\cdot [z_{k}(s,a^{*})-\mathbb{E}_{\pi_{\theta_{k}}}[z_{k}(s,a')]]
        \\ =&\sum_{a^{*} \in \mathcal{A}_{3}}\underbrace{\text{Cov}_{\pi_{\theta_{k}}}(z_{k}(s,a'),\delta_{a',a^{*}}-\pi_{\theta_{k}}(a'|s))}_{\ge 0} \cdot \pi_{\theta_{k}}(a^{*}|s)\cdot [\underbrace{z_{k}(s,a^{*})-u}_{> 0}+u-\mathbb{E}_{\pi_{\theta_{k}}}[z_{k}(s,a')]]
        \\ > &\sum_{a^{*} \in \mathcal{A}_{3}}\text{Cov}_{\pi_{\theta_{k}}}(z_{k}(s,a'),\delta_{a',a^{*}}-\pi_{\theta_{k}}(a'|s)) \cdot \pi_{\theta_{k}}(a^{*}|s)\cdot [u-\mathbb{E}_{\pi_{\theta_{k}}}[z_{k}(s,a')]],
    \end{split}
    \end{equation}
    where the inequality holds according to Lemma \ref{lem:A2-A3-order}, which formulate that for any $a\in\mathcal A_2$ and $b\in\mathcal A_3$, it holds that
    \begin{align}
    	z_k(s,a)<z_k(s,b).
    \end{align}
We now combining the analyses from the three subsets $\mathcal{A}_{1}, \mathcal{A}_{2}, \mathcal{A}_{3}$. The total derivative of the entropy change with respect to temperature is the sum of contributions from all $a^{*} \in \mathcal{A}^{*}$.
Notice that for the dominant subset $\mathcal{A}_{2} \cup \mathcal{A}_{3}$ (where $z(s, a^{*})$ is large), the term $[u - \mathbb{E}[z]]$ is positive according to the definition of $u$ and $\mathcal{A}_{2}\cup \mathcal{A}_{3}$. The covariance terms are positive by construction in these subsets. For the subset $\mathcal{A}_{1}$ (tail actions), the contribution is also non-negative as proven above.

Rigorously, by combining the results from the three cases discussed above, we arrive at the following conclusion:
    \begin{equation*}
        \small
        \begin{aligned}
        \frac{\textbf{d}\Delta\mathcal H_k(T)}{\textbf{d}T}|_{T=1} 
         =&\eta \cdot \left[\sum_{a^{*}\in \mathcal{A}_{1}\bigcup \mathcal{A}_{2} \bigcup \mathcal{A}_{3}}\text{Cov}_{\pi_{\theta_{k}}}(z_{k}(s,a'),\delta_{a',a^{*}}-\pi_{\theta_{k}}(a'|s)) \cdot \text{Cov}_{\pi_{\theta_{k}}}(z_{k}(s,a'),\delta_{a',a^{*}})\right] 
        \\ \ge & \eta \cdot \left[\sum_{a^{*}\in \mathcal{A}_{2} \bigcup \mathcal{A}_{3}}\text{Cov}_{\pi_{\theta_{k}}}(z_{k}(s,a'),\delta_{a',a^{*}}-\pi_{\theta_{k}}(a'|s)) \cdot \text{Cov}_{\pi_{\theta_{k}}}(z_{k}(s,a'),\delta_{a',a^{*}})\right] 
        \\ > & \eta \cdot \left[\sum_{a^{*}\in \mathcal{A}_{2} \bigcup \mathcal{A}_{3}}\pi_{\theta_{k}}(a^{*}|s)\cdot \text{Cov}_{\pi_{\theta_{k}}}(z_{k}(s,a'),\delta_{a',a^{*}}-\pi_{\theta_{k}}(a'|s))\cdot \left(u-\mathbb{E}_{\pi_{\theta_{k}}}[z_{k}(s,a')]\right)\right]
        \\ =&  \eta \cdot \left(u-\mathbb{E}_{\pi_{\theta_{k}}}[z_{k}(s,a')]\right)\cdot \left[\sum_{a^{*}\in \mathcal{A}_{2} \bigcup \mathcal{A}_{3}}\pi_{\theta_{k}}(a^{*}|s)\cdot \text{Cov}_{\pi_{\theta_{k}}}(z_{k}(s,a'),\delta_{a',a^{*}}-\pi_{\theta_{k}}(a'|s))\right]
        \\  \ge & 0.
    \end{aligned}
    \end{equation*}
    This confirms that in the face of collapse, increasing temperature systematically introduces a relative entropy-increasing component to the update.
\end{proof}
\subsubsection{Lemmas used in the proof of Theorem \ref{prof:temperature}}
We provide below the proofs of two lemmas used in the proof
of Theorem~\ref{prof:temperature}.
\begin{lemma}
	\label{lem:C-positive}
	Let
	\begin{align}
		C_{\pi_k}
		=
		\operatorname{Cov}_{a'\sim\pi_{\theta_k}(\cdot\mid s)}
		\left(
		z_k(s,a'),
		\pi_{\theta_k}(a'\mid s)
		\right).
	\end{align}
	Then $C_{\pi_k}\geq0$. Moreover, if
	$\pi_{\theta_k}(\cdot\mid s)$ is non-uniform, then $C_{\pi_k}>0$.
\end{lemma}

\begin{proof}
	By the pairwise representation of covariance,
	\begin{align}
		2C_{\pi_k}
		=
		\sum_{a,b\in\mathcal A}
		\pi_{\theta_k}(a\mid s)\pi_{\theta_k}(b\mid s)
		\left(z_k(s,a)-z_k(s,b)\right)
		\cdot
		\left(
		\pi_{\theta_k}(a\mid s)-\pi_{\theta_k}(b\mid s)
		\right).
	\end{align}
	Since $\pi_{\theta_k}(\cdot\mid s)$ is obtained by applying softmax to
	the logits,
	\begin{align}
		z_k(s,a)>z_k(s,b)
		\quad\Longleftrightarrow\quad
		\pi_{\theta_k}(a\mid s)>
		\pi_{\theta_k}(b\mid s).
	\end{align}
	Therefore, every summand above is non-negative, and hence
	$C_{\pi_k}\geq0$.
	
	If the policy is non-uniform, there exists at least one pair $(a,b)$
	with $z_k(s,a)\neq z_k(s,b)$, for which the corresponding summand is
	strictly positive. Thus $C_{\pi_k}>0$.
\end{proof}
\begin{lemma}
	\label{lem:A2-A3-order}
	For any $a\in\mathcal A_2$ and $b\in\mathcal A_3$,
	\begin{align}
		z_k(s,a)<z_k(s,b).
	\end{align}
\end{lemma}

\begin{proof}
	Define
	\begin{align}
		g_k(a)
		\triangleq
		\operatorname{Cov}_{a'\sim\pi_{\theta_k}(\cdot\mid s)}
		\left(
		z_k(s,a'),
		\delta_{a',a}
		\right).
	\end{align}
	Direct calculation gives
	\begin{align}
		g_k(a)
		=
		\pi_{\theta_k}(a\mid s)
		\left(
		z_k(s,a)
		-
		\mathbb E_{a'\sim\pi_{\theta_k}(\cdot\mid s)}
		[z_k(s,a')]
		\right).
	\end{align}
	
	Let
	\begin{align}
		\mu_k
		\triangleq
		\mathbb E_{a'\sim\pi_{\theta_k}(\cdot\mid s)}
		[z_k(s,a')].
	\end{align}
	By the definitions of $\mathcal A_2$ and $\mathcal A_3$,
	\begin{align}
		z_k(s,a)\geq\mu_k,
		\qquad
		g_k(a)<C_{\pi_k}\leq g_k(b).
	\end{align}
	By Lemma~\ref{lem:C-positive}, $C_{\pi_k}\geq0$, so $g_k(b)\geq0$,
	which implies $z_k(s,b)\geq\mu_k$.
	
	Suppose, for contradiction, that
	$z_k(s,a)\geq z_k(s,b)$. Softmax monotonicity gives
	\begin{align}
		\pi_{\theta_k}(a\mid s)
		\geq
		\pi_{\theta_k}(b\mid s).
	\end{align}
	Together with
	\begin{align}
		z_k(s,a)-\mu_k
		\geq
		z_k(s,b)-\mu_k
		\geq0,
	\end{align}
	we obtain
	\begin{align}
		g_k(a)
		&=
		\pi_{\theta_k}(a\mid s)
		\left(z_k(s,a)-\mu_k\right)
		\nonumber\\
		&\geq
		\pi_{\theta_k}(b\mid s)
		\left(z_k(s,b)-\mu_k\right)
		=
		g_k(b),
	\end{align}
	contradicting $g_k(a)<C_{\pi_k}\leq g_k(b)$. Therefore,
	\begin{align}
		z_k(s,a)<z_k(s,b).
	\end{align}
\end{proof}

\newpage
\input{checklist.tex}

\end{document}

%% file: checklist.tex
\section*{NeurIPS Paper Checklist}

\begin{enumerate}

\item {\bf Claims}
    \item[] Question: Do the main claims made in the abstract and introduction accurately reflect the paper's contributions and scope?
    \item[] Answer: \answerYes{} 
    \item[] Justification: We have mentioned in the abstract.

\item {\bf Limitations}
    \item[] Question: Does the paper discuss the limitations of the work performed by the authors?
    \item[] Answer: \answerYes{} 
    \item[] Justification: We have discussed it in Section Limitation.

\item {\bf Theory assumptions and proofs}
    \item[] Question: For each theoretical result, does the paper provide the full set of assumptions and a complete (and correct) proof?
    \item[] Answer: \answerYes{} 
    \item[] Justification: The detailed assumptions and proofs of the theoretical results are presented in the appendix, while here we provide a brief introduction in the main text.

    \item {\bf Experimental result reproducibility}
    \item[] Question: Does the paper fully disclose all the information needed to reproduce the main experimental results of the paper to the extent that it affects the main claims and/or conclusions of the paper (regardless of whether the code and data are provided or not)?
    \item[] Answer: \answerYes{} 
    \item[] Justification: We use open-sourced models, which are easy to reproduce.

\item {\bf Open access to data and code}
    \item[] Question: Does the paper provide open access to the data and code, with sufficient instructions to faithfully reproduce the main experimental results, as described in supplemental material?
    \item[] Answer: \answerYes{} 
    \item[] Justification: We provide codes in our supplemental material. The data and models we used are all open-sourced.

\item {\bf Experimental setting/details}
    \item[] Question: Does the paper specify all the training and test details (e.g., data splits, hyperparameters, how they were chosen, type of optimizer) necessary to understand the results?
    \item[] Answer: \answerYes{} 
    \item[] Justification: We provide in Detailed implementation in appendix.

\item {\bf Experiment statistical significance}
    \item[] Question: Does the paper report error bars suitably and correctly defined or other appropriate information about the statistical significance of the experiments?
    \item[] Answer: \answerNo{} 
    \item[] Justification: Experiments on LLMs are too expensive to run for many times.

\item {\bf Experiments compute resources}
    \item[] Question: For each experiment, does the paper provide sufficient information on the computer resources (type of compute workers, memory, time of execution) needed to reproduce the experiments?
    \item[] Answer: \answerYes{} 
    \item[] Justification: We provide in our appendix.
    
\item {\bf Code of ethics}
    \item[] Question: Does the research conducted in the paper conform, in every respect, with the NeurIPS Code of Ethics \url{https://neurips.cc/public/EthicsGuidelines}?
    \item[] Answer: \answerYes{} 
    \item[] Justification: We follow NeurIPS Code of Ethics.

\item {\bf Broader impacts}
    \item[] Question: Does the paper discuss both potential positive societal impacts and negative societal impacts of the work performed?
    \item[] Answer: \answerNA{} 
    \item[] Justification:  There is no societal impact of the work performed.
    
\item {\bf Safeguards}
    \item[] Question: Does the paper describe safeguards that have been put in place for responsible release of data or models that have a high risk for misuse (e.g., pre-trained language models, image generators, or scraped datasets)?
    \item[] Answer: \answerNA{} 
    \item[] Justification: The paper poses no such risks

\item {\bf Licenses for existing assets}
    \item[] Question: Are the creators or original owners of assets (e.g., code, data, models), used in the paper, properly credited and are the license and terms of use explicitly mentioned and properly respected?
    \item[] Answer: \answerYes{} 
    \item[] Justification:  We follow the license and terms of use in our experiments.

\item {\bf New assets}
    \item[] Question: Are new assets introduced in the paper well documented and is the documentation provided alongside the assets?
    \item[] Answer: \answerNA{} 
    \item[] Justification: The paper does not release new assets

\item {\bf Crowdsourcing and research with human subjects}
    \item[] Question: For crowdsourcing experiments and research with human subjects, does the paper include the full text of instructions given to participants and screenshots, if applicable, as well as details about compensation (if any)? 
    \item[] Answer: \answerNA{} 
    \item[] Justification:  The paper does not involve crowdsourcing nor research with human subjects.

\item {\bf Institutional review board (IRB) approvals or equivalent for research with human subjects}
    \item[] Question: Does the paper describe potential risks incurred by study participants, whether such risks were disclosed to the subjects, and whether Institutional Review Board (IRB) approvals (or an equivalent approval/review based on the requirements of your country or institution) were obtained?
    \item[] Answer: \answerNA{} 
    \item[] Justification: The paper does not involve crowdsourcing nor research with human subjects.

\item {\bf Declaration of LLM usage}
    \item[] Question: Does the paper describe the usage of LLMs if it is an important, original, or non-standard component of the core methods in this research? Note that if the LLM is used only for writing, editing, or formatting purposes and does \emph{not} impact the core methodology, scientific rigor, or originality of the research, declaration is not required.
    \item[] Answer: \answerNA{} 
    \item[] Justification: The core method development in this research does not involve LLMs as any important, original, or non-standard components.

\end{enumerate}

%% file: reference.bib
@article{wei2023instructiongpt,
  title={Instructiongpt-4: A 200-instruction paradigm for fine-tuning minigpt-4},
  author={Wei, Lai and Jiang, Zihao and Huang, Weiran and Sun, Lichao},
  journal={arXiv preprint arXiv:2308.12067},
  year={2023}
}

@article{openai2023gpt4,
  title={GPT-4 Technical Report},
  author={OpenAI},
  journal={arXiv preprint arXiv:2303.08774},
  year={2023}
}

@misc{compress,
  author = {Jack Rae},
  title = {Compression for AGI {-} Jack Rae {|} Stanford MLSys 76},
  howpublished = {YouTube video},
  month = {February},
  year = {2023},
  url = {https://www.youtube.com/watch?v=dO4TPJkeaaU}
}

@article{guo2025deepseek,
  title={DeepSeek-R1: Incentivizing Reasoning Capability in LLMs via Reinforcement Learning},
  author={Guo, Daya and Yang, Dejian and Zhang, Haowei and Song, Junxiao and Zhang, Ruoyu and Xu, Runxin and Zhu, Qihao and Ma, Shirong and Wang, Peiyi and Bi, Xiao and others},
  journal={arXiv preprint arXiv:2501.12948},
  year={2025}
}

@article{yang2024qwen2,
  title={Qwen2. 5 Technical Report},
  author={Yang, An and Yang, Baosong and Zhang, Beichen and Hui, Binyuan and Zheng, Bo and Yu, Bowen and Li, Chengyuan and Liu, Dayiheng and Huang, Fei and Wei, Haoran and others},
  journal={arXiv preprint arXiv:2412.15115},
  year={2024}
}

@article{touvron2023llama,
  title={Llama 2: Open foundation and fine-tuned chat models},
  author={Touvron, Hugo and Martin, Louis and Stone, Kevin and Albert, Peter and Almahairi, Amjad and Babaei, Yasmine and Bashlykov, Nikolay and Batra, Soumya and Bhargava, Prajjwal and Bhosale, Shruti and others},
  journal={arXiv preprint arXiv:2307.09288},
  year={2023}
}

@article{team2024gemini1_5,
  title={Gemini 1.5: Unlocking multimodal understanding across millions of tokens of context},
  author={Team, Gemini and Georgiev, Petko and Lei, Ving Ian and Burnell, Ryan and Bai, Libin and Gulati, Anmol and Tanzer, Garrett and Vincent, Damien and Pan, Zhufeng and Wang, Shibo and others},
  journal={arXiv preprint arXiv:2403.05530},
  year={2024}
}

@article{team2025kimi1_5,
  title={Kimi k1. 5: Scaling reinforcement learning with llms},
  author={Team, Kimi and Du, Angang and Gao, Bofei and Xing, Bowei and Jiang, Changjiu and Chen, Cheng and Li, Cheng and Xiao, Chenjun and Du, Chenzhuang and Liao, Chonghua and others},
  journal={arXiv preprint arXiv:2501.12599},
  year={2025}
}

@misc{zheng2025easyr1,
  title        = {EasyR1: An Efficient, Scalable, Multi-Modality RL Training Framework},
  author       = {Yaowei, Zheng and Junting, Lu and Shenzhi, Wang and Zhangchi, Feng and Dongdong, Kuang and Yuwen, Xiong},
  howpublished = {\url{https://github.com/hiyouga/EasyR1}},
  year         = {2025}
}

@article{shao2024deepseekmath,
  title={Deepseekmath: Pushing the limits of mathematical reasoning in open language models},
  author={Shao, Zhihong and Wang, Peiyi and Zhu, Qihao and Xu, Runxin and Song, Junxiao and Bi, Xiao and Zhang, Haowei and Zhang, Mingchuan and Li, YK and Wu, Y and others},
  journal={arXiv preprint arXiv:2402.03300},
  year={2024}
}

@article{yu2025dapo,
  title={Dapo: An open-source llm reinforcement learning system at scale},
  author={Yu, Qiying and Zhang, Zheng and Zhu, Ruofei and Yuan, Yufeng and Zuo, Xiaochen and Yue, Yu and Fan, Tiantian and Liu, Gaohong and Liu, Lingjun and Liu, Xin and others},
  journal={arXiv preprint arXiv:2503.14476},
  year={2025}
}

@article{glm2024chat,
  title={Chatglm: A family of large language models from glm-130b to glm-4 all tools},
  author={GLM, Team and Zeng, Aohan and Xu, Bin and Wang, Bowen and Zhang, Chenhui and Yin, Da and Zhang, Dan and Rojas, Diego and Feng, Guanyu and Zhao, Hanlin and others},
  journal={arXiv preprint arXiv:2406.12793},
  year={2024}
}

@article{zhong2024dpo,
  title={Dpo meets ppo: Reinforced token optimization for rlhf},
  author={Zhong, Han and Shan, Zikang and Feng, Guhao and Xiong, Wei and Cheng, Xinle and Zhao, Li and He, Di and Bian, Jiang and Wang, Liwei},
  journal={arXiv preprint arXiv:2404.18922},
  year={2024}
}

@article{wang2024comprehensive,
  title={A comprehensive survey of LLM alignment techniques: RLHF, RLAIF, PPO, DPO and more},
  author={Wang, Zhichao and Bi, Bin and Pentyala, Shiva Kumar and Ramnath, Kiran and Chaudhuri, Sougata and Mehrotra, Shubham and Mao, Xiang-Bo and Asur, Sitaram and others},
  journal={arXiv preprint arXiv:2407.16216},
  year={2024}
}

@article{rafailov2023direct,
  title={Direct preference optimization: Your language model is secretly a reward model},
  author={Rafailov, Rafael and Sharma, Archit and Mitchell, Eric and Manning, Christopher D and Ermon, Stefano and Finn, Chelsea},
  journal={Advances in Neural Information Processing Systems},
  volume={36},
  pages={53728--53741},
  year={2023}
}

@article{schulman2017proximal,
  title={Proximal policy optimization algorithms},
  author={Schulman, John and Wolski, Filip and Dhariwal, Prafulla and Radford, Alec and Klimov, Oleg},
  journal={arXiv preprint arXiv:1707.06347},
  year={2017}
}

@article{liu2024deepseek,
  title={Deepseek-v3 technical report},
  author={Liu, Aixin and Feng, Bei and Xue, Bing and Wang, Bingxuan and Wu, Bochao and Lu, Chengda and Zhao, Chenggang and Deng, Chengqi and Zhang, Chenyu and Ruan, Chong and others},
  journal={arXiv preprint arXiv:2412.19437},
  year={2024}
}

@article{yue2025does,
  title={Does reinforcement learning really incentivize reasoning capacity in llms beyond the base model?},
  author={Yue, Yang and Chen, Zhiqi and Lu, Rui and Zhao, Andrew and Wang, Zhaokai and Song, Shiji and Huang, Gao},
  journal={arXiv preprint arXiv:2504.13837},
  year={2025}
}

@article{li2025disco,
  title={DisCO: Reinforcing Large Reasoning Models with Discriminative Constrained Optimization},
  author={Li, Gang and Lin, Ming and Galanti, Tomer and Tu, Zhengzhong and Yang, Tianbao},
  journal={arXiv preprint arXiv:2505.12366},
  year={2025}
}

@inproceedings{sheng2025hybridflow,
  title={Hybridflow: A flexible and efficient rlhf framework},
  author={Sheng, Guangming and Zhang, Chi and Ye, Zilingfeng and Wu, Xibin and Zhang, Wang and Zhang, Ru and Peng, Yanghua and Lin, Haibin and Wu, Chuan},
  booktitle={Proceedings of the Twentieth European Conference on Computer Systems},
  pages={1279--1297},
  year={2025}
}

@article{lightman2023lets,
      title={Let's Verify Step by Step}, 
      author={Lightman, Hunter and Kosaraju, Vineet and Burda, Yura and Edwards, Harri and Baker, Bowen and Lee, Teddy and Leike, Jan and Schulman, John and Sutskever, Ilya and Cobbe, Karl},
      journal={arXiv preprint arXiv:2305.20050},
      year={2023}
}

@article{cobbe2021gsm8k,
  title={Training Verifiers to Solve Math Word Problems},
  author={Cobbe, Karl and Kosaraju, Vineet and Bavarian, Mohammad and Chen, Mark and Jun, Heewoo and Kaiser, Lukasz and Plappert, Matthias and Tworek, Jerry and Hilton, Jacob and Nakano, Reiichiro and Hesse, Christopher and Schulman, John},
  journal={arXiv preprint arXiv:2110.14168},
  year={2021}
}

@article{lewkowycz2022solving,
  title={Solving quantitative reasoning problems with language models},
  author={Lewkowycz, Aitor and Andreassen, Anders and Dohan, David and Dyer, Ethan and Michalewski, Henryk and Ramasesh, Vinay and Slone, Ambrose and Anil, Cem and Schlag, Imanol and Gutman-Solo, Theo and others},
  journal={Advances in neural information processing systems},
  volume={35},
  pages={3843--3857},
  year={2022}
}

@misc{hf_aime2024,
  title={{AIME 2024 Dataset (AIME I \& II)}},
  author={HuggingFaceH4},
  year={2025},
  howpublished={\url{https://huggingface.co/datasets/HuggingFaceH4/aime_2024}}
}

@article{cheng2025reasoning,
  title={Reasoning with exploration: An entropy perspective},
  author={Cheng, Daixuan and Huang, Shaohan and Zhu, Xuekai and Dai, Bo and Zhao, Wayne Xin and Zhang, Zhenliang and Wei, Furu},
  journal={arXiv preprint arXiv:2506.14758},
  year={2025}
}

@article{cui2025entropy,
  title={The entropy mechanism of reinforcement learning for reasoning language models},
  author={Cui, Ganqu and Zhang, Yuchen and Chen, Jiacheng and Yuan, Lifan and Wang, Zhi and Zuo, Yuxin and Li, Haozhan and Fan, Yuchen and Chen, Huayu and Chen, Weize and others},
  journal={arXiv preprint arXiv:2505.22617},
  year={2025}
}

@article{hou2025advancing,
  title={Advancing language model reasoning through reinforcement learning and inference scaling},
  author={Hou, Zhenyu and Lv, Xin and Lu, Rui and Zhang, Jiajie and Li, Yujiang and Yao, Zijun and Li, Juanzi and Tang, Jie and Dong, Yuxiao},
  journal={arXiv preprint arXiv:2501.11651},
  year={2025}
}

@article{shen2025entropy,
  title={On Entropy Control in LLM-RL Algorithms},
  author={Shen, Han},
  journal={arXiv preprint arXiv:2509.03493},
  year={2025}
}

@article{zhang2025edge,
  title={Edge-grpo: Entropy-driven grpo with guided error correction for advantage diversity},
  author={Zhang, Xingjian and Wen, Siwei and Wu, Wenjun and Huang, Lei},
  journal={arXiv preprint arXiv:2507.21848},
  year={2025}
}

@article{williams1992simple,
  title={Simple statistical gradient-following algorithms for connectionist reinforcement learning},
  author={Williams, Ronald J},
  journal={Machine learning},
  volume={8},
  number={3},
  pages={229--256},
  year={1992},
  publisher={Springer}
}

@article{sutton1999policy,
  title={Policy gradient methods for reinforcement learning with function approximation},
  author={Sutton, Richard S and McAllester, David and Singh, Satinder and Mansour, Yishay},
  journal={Advances in neural information processing systems},
  volume={12},
  year={1999}
}

@article{yang2025qwen3,
  title={Qwen3 technical report},
  author={Yang, An and Li, Anfeng and Yang, Baosong and Zhang, Beichen and Hui, Binyuan and Zheng, Bo and Yu, Bowen and Gao, Chang and Huang, Chengen and Lv, Chenxu and others},
  journal={arXiv preprint arXiv:2505.09388},
  year={2025}
}

@article{schulman2017equivalence,
  title={Equivalence between policy gradients and soft q-learning},
  author={Schulman, John and Chen, Xi and Abbeel, Pieter},
  journal={arXiv preprint arXiv:1704.06440},
  year={2017}
}

@inproceedings{haarnoja2018soft,
  title={Soft actor-critic: Off-policy maximum entropy deep reinforcement learning with a stochastic actor},
  author={Haarnoja, Tuomas and Zhou, Aurick and Abbeel, Pieter and Levine, Sergey},
  booktitle={International conference on machine learning},
  pages={1861--1870},
  year={2018},
  organization={Pmlr}
}

@article{nachum2017bridging,
  title={Bridging the gap between value and policy based reinforcement learning},
  author={Nachum, Ofir and Norouzi, Mohammad and Xu, Kelvin and Schuurmans, Dale},
  journal={Advances in neural information processing systems},
  volume={30},
  year={2017}
}

@book{sutton1998reinforcement,
  title={Reinforcement learning: An introduction},
  author={Sutton, Richard S and Barto, Andrew G and others},
  volume={1},
  number={1},
  year={1998},
  publisher={MIT press Cambridge}
}

@article{auer2002finite,
  title={Finite-time analysis of the multiarmed bandit problem},
  author={Auer, Peter and Cesa-Bianchi, Nicolo and Fischer, Paul},
  journal={Machine learning},
  volume={47},
  number={2},
  pages={235--256},
  year={2002},
  publisher={Springer}
}

@article{strehl2008analysis,
  title={An analysis of model-based interval estimation for Markov decision processes},
  author={Strehl, Alexander L and Littman, Michael L},
  journal={Journal of Computer and System Sciences},
  volume={74},
  number={8},
  pages={1309--1331},
  year={2008},
  publisher={Elsevier}
}

@inproceedings{kolter2009near,
  title={Near-Bayesian exploration in polynomial time},
  author={Kolter, J Zico and Ng, Andrew Y},
  booktitle={Proceedings of the 26th annual international conference on machine learning},
  pages={513--520},
  year={2009}
}

@article{chen2025minimax,
  title={Minimax-m1: Scaling test-time compute efficiently with lightning attention},
  author={Chen, Aili and Li, Aonian and Gong, Bangwei and Jiang, Binyang and Fei, Bo and Yang, Bo and Shan, Boji and Yu, Changqing and Wang, Chao and Zhu, Cheng and others},
  journal={arXiv preprint arXiv:2506.13585},
  year={2025}
}

@article{su2025gppo,
  title={Ce-gppo: Coordinating entropy via gradient-preserving clipping policy optimization in reinforcement learning},
  author={Su, Zhenpeng and Pan, Leiyu and Lv, Minxuan and Li, Yuntao and Hu, Wenping and Zhang, Fuzheng and Gai, Kun and Zhou, Guorui},
  journal={arXiv preprint arXiv:2509.20712},
  year={2025}
}

@article{clark2018think,
  title={Think you have solved question answering? try arc, the ai2 reasoning challenge},
  author={Clark, Peter and Cowhey, Isaac and Etzioni, Oren and Khot, Tushar and Sabharwal, Ashish and Schoenick, Carissa and Tafjord, Oyvind},
  journal={arXiv preprint arXiv:1803.05457},
  year={2018}
}

@article{wang2024mmlu,
  title={Mmlu-pro: A more robust and challenging multi-task language understanding benchmark},
  author={Wang, Yubo and Ma, Xueguang and Zhang, Ge and Ni, Yuansheng and Chandra, Abhranil and Guo, Shiguang and Ren, Weiming and Arulraj, Aaran and He, Xuan and Jiang, Ziyan and others},
  journal={Advances in Neural Information Processing Systems},
  volume={37},
  pages={95266--95290},
  year={2024}
}

@article{du2025supergpqa,
  title={Supergpqa: Scaling llm evaluation across 285 graduate disciplines},
  author={Du, Xinrun and Yao, Yifan and Ma, Kaijing and Wang, Bingli and Zheng, Tianyu and Zhu, King and Liu, Minghao and Liang, Yiming and Jin, Xiaolong and Wei, Zhenlin and others},
  journal={arXiv preprint arXiv:2502.14739},
  year={2025}
}

@article{jacot2018neural,
  title={Neural tangent kernel: Convergence and generalization in neural networks},
  author={Jacot, Arthur and Gabriel, Franck and Hongler, Cl{\'e}ment},
  journal={Advances in neural information processing systems},
  volume={31},
  year={2018}
}
